\documentclass{article}
\usepackage{microtype}
\usepackage{graphicx}
\usepackage{subcaption}
\usepackage{multirow} 
\usepackage{booktabs} 
\usepackage{hyperref}

\usepackage[preprint]{icml2026}

\usepackage{amsmath}
\usepackage{amssymb}
\usepackage{mathtools}
\usepackage{amsthm}
\usepackage{float} 

\usepackage{xargs}   
\usepackage{ifthen}  
\usepackage{bm}      
\usepackage{xcolor}  

\newcommand{\tbar}{\mathrel{\raisebox{0.15ex}{$\scriptscriptstyle\mid$}}}
\def\eqdef{\vcentcolon=}
\def\gauss{\mathcal{N}}
\def\gausspdf{\mathrm{N}}
\def\rmd{\mathrm{d}}
\def\eqsp{\,}
\def\obs{\vect{y}}
\def\operator{\mathcal{A}}
\def\dimobs{{d'}}

\def\Id{\mat{I}}
\def\pE{\mathbb{E}}

\newcommandx\fw[4][4=]{
        \ifthenelse{\equal{#3}{}}{
            q^{#4} _{\smash{#1}}(\cdot|#2)
        }{
            q^{#4} _{\smash{#1}}(#3|#2)
        }
    }

\newcommandx\pdata[4][4=]{
    \ifthenelse{\equal{#2}{}}{
        \ifthenelse{\equal{#3}{}}{
            p^{#4} _{#1}
        }{
            p^{#4} _{#1}(#3)
        }
    }{
        \ifthenelse{\equal{#3}{}}{
            p^{#4} _{#1}(\cdot|#2)
        }{
            p^{#4} _{#1}(#3|#2)
        }
    }}

\newcommandx\posterior[4][4=]{
    \ifthenelse{\equal{#2}{}}{
        \ifthenelse{\equal{#3}{}}{
            \pi^{#4} _{#1}(\cdot | \obs, \operator)
        }{
            \pi^{#4} _{#1}(#3 | \obs, \operator)
        }
    }{
        \ifthenelse{\equal{#3}{}}{
            \pi^{#4} _{#1}(\cdot|#2, \obs, \operator)
        }{
            \pi^{#4} _{#1}(#3|#2, \obs, \operator)
        }
    }}

\newcommandx\approxposterior[4][4=]{
    \ifthenelse{\equal{#2}{}}{
        \ifthenelse{\equal{#3}{}}{
            \hat{\pi}^{#4} _{#1}(\cdot | \obs, \operator)
        }{
            \hat{\pi}^{#4} _{#1}(#3 | \obs, \operator)
        }
    }{
        \ifthenelse{\equal{#3}{}}{
            \hat{\pi}^{#4} _{#1}(\cdot|#2, \obs, \operator)
        }{
            \hat{\pi}^{#4} _{#1}(#3|#2, \obs, \operator)
        }
    }}

\newcommandx\barposterior[4][4=]{
    \ifthenelse{\equal{#2}{}}{
        \ifthenelse{\equal{#3}{}}{
            \bar{\pi}^{#4} _{#1}(\cdot | \obs, \operator)
        }{
            \bar{\pi}^{#4} _{#1}(#3 | \obs, \operator)
        }
    }{
        \ifthenelse{\equal{#3}{}}{
            \bar{\pi}^{#4} _{#1}(\cdot|#2, \obs, \operator)
        }{
            \bar{\pi}^{#4} _{#1}(#3|#2, \obs, \operator)
        }
    }}

\newcommandx\denoiser[4][4=]{
    \ifthenelse{\equal{#2}{}}{
        \ifthenelse{\equal{#3}{}}{
            D^{#4}_{0\tbar #1}
        }{
            D^{#4} _{0\tbar #1}(#3)
        }
    }{
        \ifthenelse{\equal{#3}{}}{
            D^{#4} _{0\tbar #1}(\cdot|#2)
        }{
            D^{#4} _{0\tbar #1}(#3|#2)
        }
    }
}

\newcommand\lklh[2]{
    \ifthenelse{\equal{#2}{}}{
        \ell_{#1}(\obs|\cdot)
    }{
        \ell_{#1}(\obs|#2)
    }}

\newcommand\lklhwithoutobs[2]{
    \ifthenelse{\equal{#2}{}}{
        \ell_{#1}(\cdot|\cdot)
    }{
        \ell_{#1}(\cdot|#2)
    }}

\newcommandx\hlklh[3][3=]{
    \ifthenelse{\equal{#2}{}}{
        \hat{\ell}^{#3} _{#1}(\obs|\cdot)
    }{
        \hat{\ell}^{#3} _{#1}(\obs|#2)
    }}

\newcommand{\KL}{D_{\mathrm{KL}}}

\newcommand{\KLdiv}[2]{\KL\left(#1 \,\|\, #2\right)}

\newcommand{\oned}[1]{\vect{1}_{#1}}

\newcommand{\context}{\left(\vect{x}_0, \vect{x}_t, s, t, \obs, \operator\right)}

\newcommand{\ffhq}{\texttt{FFHQ64}}

\newcommand\vect[1]{\bm{#1}}
\newcommand\rdmvect[1]{\bm{#1}}
\newcommand\mat[1]{\bm{#1}}

\usepackage{enumitem}
\usepackage{xspace}
\newcommand{\ours}{LAVPS\xspace}
\newcommand{\ourmethod}{Likelihood-guided Amortized Variational Posterior Sampling (LAVPS)\xspace}
\newcommand{\RETURN}{\STATE \textbf{return }}

\usepackage[capitalize,noabbrev]{cleveref}

\theoremstyle{plain}

\theoremstyle{definition}

\theoremstyle{remark}

\usepackage[textsize=tiny]{todonotes}

\icmltitlerunning{Fast and Robust Likelihood-Guided Diffusion Posterior Sampling with Amortized Variational Inference}

\begin{document}

\twocolumn[
  \icmltitle{Fast and Robust Likelihood-Guided Diffusion Posterior Sampling with Amortized Variational Inference}

  \icmlsetsymbol{equal}{*}

  \begin{icmlauthorlist}
    \icmlauthor{Léon Zheng}{comp}
    \icmlauthor{Thomas Hirtz}{comp}
    \icmlauthor{Yazid Janati}{ifm}
    \icmlauthor{Eric Moulines}{sch}
  \end{icmlauthorlist}

  \icmlaffiliation{ifm}{Institute of Foundation Models, MBZUAI}
  \icmlaffiliation{comp}{Huawei Lagrange Mathematics and Computing Research Center, Paris, France}
  \icmlaffiliation{sch}{Mohamed Bin Zayed University of AI, Chair of the ML Department, EPITA, Laboratoire Recherche de l'EPITA}

  \icmlcorrespondingauthor{Léon Zheng}{leon.zheng@polytechnique.org}
  \icmlkeywords{Machine Learning, ICML}

  \vskip 0.3in
]

\printAffiliationsAndNotice{} 

\begin{abstract}
    Zero-shot diffusion posterior sampling offers a flexible framework for inverse problems by accommodating arbitrary degradation operators at test time, but incurs high computational cost due to repeated likelihood-guided updates. In contrast, previous amortized diffusion approaches enable fast inference by replacing likelihood-based sampling with implicit inference models, but at the expense of robustness to unseen degradations. We introduce an amortization strategy for diffusion posterior sampling that preserves explicit likelihood guidance by amortizing the inner optimization problems arising in variational diffusion posterior sampling. This accelerates inference for in-distribution degradations while maintaining robustness to previously unseen operators, thereby improving the trade-off between efficiency and flexibility in diffusion-based inverse problems.
\end{abstract}

\section{Introduction}
An inverse problem aims to recover an unknown signal from incomplete, noisy observations. Such problems arise in a wide range of applications, including imaging tasks such as super-resolution, inpainting, and deblurring. These problems are typically difficult because they are ill-posed: a single perturbed observation may correspond to multiple plausible reconstructions.
To overcome this difficulty, recent work has employed diffusion models \citep{ho2020denoising,nichol2021improved} pre-trained on large-scale image datasets. These models serve as powerful \emph{priors} that guide the solution of inverse problems towards reconstructions that are both realistic and consistent with the observed data.

A prominent line of work is \emph{zero-shot diffusion posterior sampling}
\citep{kadkhodaie2021stochastic,song2021score,kawar2022denoising, lugmayr2022repaint, wang2023zeroshot, song2023pseudoinverse, chung2023diffusion},
which encompasses a family of methods that require \emph{no additional training} beyond a pre-trained diffusion model.
At test time, these approaches aim to sample from the \emph{posterior} distribution of reconstructions conditioned on the degraded observation, assuming that the underlying degradation process is known and can be expressed through a \emph{likelihood} probability density function.
By design, such methods are flexible and can accommodate arbitrary degradations at inference.
This stands in contrast to \emph{supervised diffusion} methods \citep{saharia2022palette,saharia2022image}, which achieve strong performance on reconstruction tasks involving degradation operators seen during training on clean–degraded pairs, but typically struggle to generalize to unseen degradations during training.

The flexibility of current zero-shot diffusion posterior sampling methods comes, however, at the cost of increased inference-time computation. This overhead stems from the need to repeatedly evaluate an approximation of the so-called \emph{likelihood guidance} \citep{chung2023diffusion, song2023pseudoinverse, rozet2024learning} at each step of the reverse diffusion process in order to steer samples toward the target posterior distribution. Different zero-shot approaches mainly differ in how this likelihood guidance term is heuristically approximated, leading to varying trade-offs between inference computational cost and reconstruction quality.

To address the high inference cost of zero-shot posterior sampling, a natural idea is to leverage amortized optimization \citep{amos2023tutorial}, which shifts computation from inference to an upstream training phase. However, existing amortized diffusion approaches \citep{lee2024diffusion,feng2023score,feng2024variational} typically replace likelihood-based posterior sampling with an implicit inference model. While this enables fast inference for degradations seen during training, it inherently ties the method to a fixed set of operators and severely limits robustness to unseen degradations.

This exposes a fundamental tension between efficiency and flexibility in diffusion-based inverse problems, and motivates the following question: \emph{Can zero-shot diffusion posterior sampling be accelerated through amortization, while preserving its ability to handle arbitrary (unseen) degradation operators at test time?}

\begin{figure*}[t] 
    \centering
    \includegraphics[width=\linewidth]{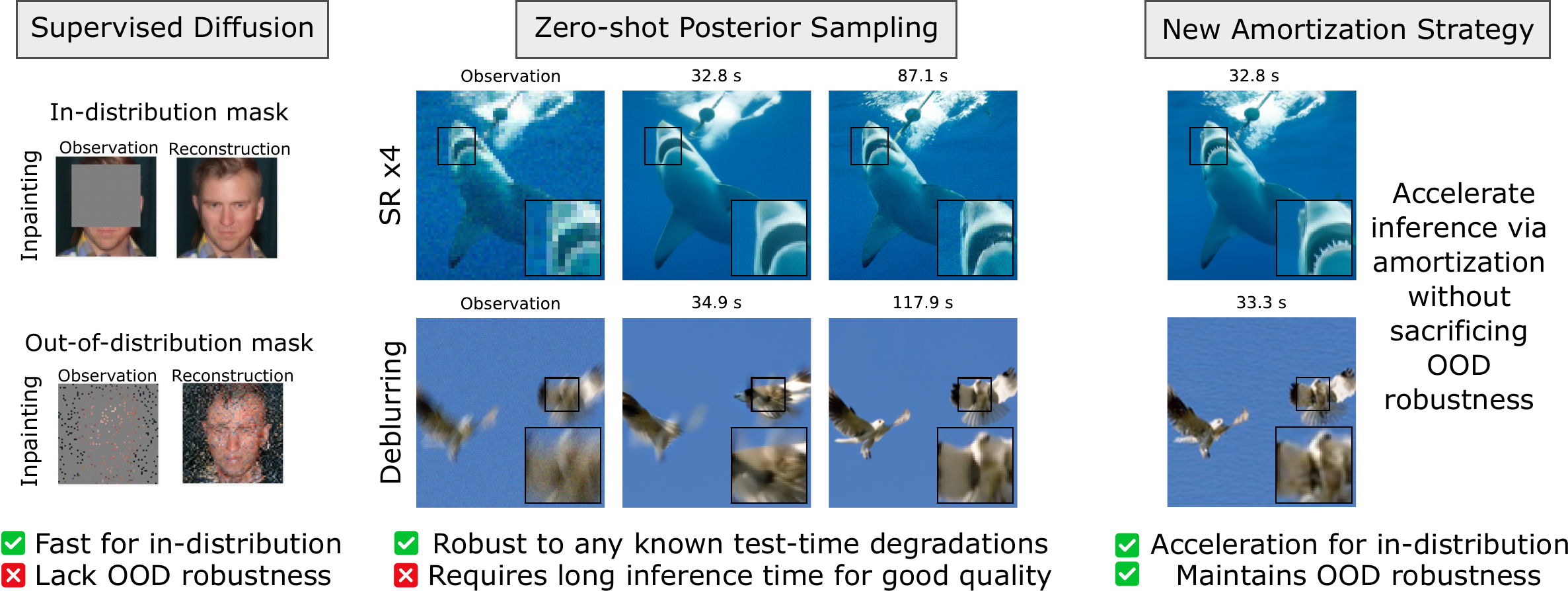}
    \caption{Summary of the novel amortization strategy. \textbf{(Left)} Illustration of the lack of robustness of supervised diffusion models to out-of-distribution operators, shown on the inpainting task using the \texttt{FFHQ} dataset (cf.~\Cref{app:lack_robustness_ood_sft}). \textbf{(Middle)} Reconstruction results for $\times4$ super-resolution and motion deblurring obtained via zero-shot posterior sampling on ImageNet, at different sampling times, highlighting the high inference cost of these methods. \textbf{(Right)} Our amortization strategy significantly accelerates inference compared to zero-shot posterior sampling (cf.~\Cref{subsec:acceleration_id}), while preserving flexibility to handle out-of-distribution operators (cf.~\Cref{subsec:robustesse_ood}).
    }
    \label{fig:figure-intro}
\end{figure*}

\paragraph{Contributions.}
We introduce a novel amortization strategy for diffusion posterior sampling that reconciles inference-time efficiency with the flexibility of zero-shot methods (\Cref{fig:figure-intro}). Rather than replacing likelihood-based posterior sampling with a fully implicit inference model, we amortize the inner optimization problems arising at each timestep of so-called \emph{variational diffusion posterior sampling} methods \citep{janati2025a,moufad2025variational} {(cf.~\Cref{sec:method} for more details)}, while preserving explicit likelihood guidance at test time. This design enables significant acceleration for degradation operators seen during training, without sacrificing the ability to handle arbitrary (and previously unseen) degradations at inference. As a result, our method, named \textbf{\ours}---standing for Likelihood-guided Amortized Variational Posterior Sampling---strictly improves over zero-shot approaches under limited inference budgets for in-distribution operators, while retaining their robustness to out-of-distribution degradations, where supervised and fully amortized diffusion methods typically fail. To the best of our knowledge, this is the first work to demonstrate that amortization and likelihood-guided diffusion can be combined to jointly achieve efficiency and robustness in diffusion-based inverse problems.

\section{Background and Motivation}
\label{sec:background}

\paragraph{Notations.} The Gaussian distribution of mean $\vect{\mu}$ and covariance $\mat{\Sigma}$ is denoted by $\gauss(\vect{\mu}, \mat{\Sigma})$. Its density is denoted as $\vect{x} \mapsto \gausspdf\left(\vect{x}; \vect{\mu}, \mat{\Sigma} \right)$. The identity matrix of size $d$ is denoted as $\Id_d$.
$\rdmvect{X} \sim p$ indicates that $\rdmvect{X}$ is a random variable whose probability distribution has density $p$.
We use $\propto$ to define a probability density up to a normalization constant.

\subsection{Diffusion Models for Inverse Problems}

\paragraph{Diffusion Models.} 
Denoising Diffusion Models (DDMs) \citep{sohl2015deep,song2019generative,ho2020denoising} define a generative process for a data distribution
$\pdata{0}{}{} \eqdef \pdata{\mathrm{data}}{}{}$,
initialized from a non-informative base distribution
$\pdata{T}{}{} \eqdef \gauss(0,\Id_d)$.
The model then sequentially samples from increasingly structured intermediate distributions $\pdata{t}{}{}$ in reverse order until $\pdata{0}{}{}$ is recovered.
The probability path $(\pdata{t}{}{})_{t=0}^T$ is determined by the \emph{forward} diffusion process:
$
	\pdata{t}{}{\vect{x}_t} := \int \fw{t \tbar 0}{\vect{x}_0}{\vect{x}_t} \pdata{0}{}{\vect{x}_0} \, \rmd \vect{x}_0
$
where we define for all $s < t$, 
\begin{equation}
	\label{eq:forward-process}
	\fw{t \tbar s}{\vect{x}_s}{\vect{x}_t} := \gausspdf\left(\vect{x}_t; \alpha_{t\tbar s} \vect{x}_s, \sigma^2 _{t\tbar s} \Id_d \right),
\end{equation}
with $\alpha_{t\tbar s} \eqdef \alpha_t / \alpha_s$, $\sigma_{t\tbar s}^2  \eqdef  \sigma^2 _t - \alpha^2 _{t\tbar s} \sigma^2 _s$ where $(\alpha_s)_{s=0} ^T$, $\alpha_0 = 1$ and $(\sigma_t)_{t = 0} ^T$, $\sigma_0 = 0$ are positive sequences chosen so that the signal-to-noise ratio $(\alpha^2 _t / \sigma^2 _t)_{t > 1}$ decreases monotonically. A popular choice of schedule is $\alpha_t = 1 - t/T$ and $\sigma_t = t/T$ \citep{lipman2023flow, esser2024scaling}. 
For $s < t$, define the Gaussian transition $\fw{s\tbar 0, t}{\vect{x}_0, \vect{x}_t}{\vect{x}_s} \propto  \fw{s\tbar 0}{\vect{x}_0}{\vect{x}_s} \fw{t\tbar s}{\vect{x}_s}{\vect{x}_t}$ and 
the reverse transitions ($0 < s < t$)
\begin{equation}
	\label{eq:unconditional-ideal-backward-kernel}
	\pdata{s\tbar t}{\vect{x}_t}{\vect{x}_s} \eqdef \int \fw{s\tbar 0, t}{\vect{x}_0, \vect{x}_t}{\vect{x}_s} \pdata{0\tbar t}{\vect{x}_t}{\vect{x}_0} \, \rmd \vect{x}_0 \eqsp,
\end{equation}
which can be shown to satisfy the recursion $\pdata{s}{}{\vect{x}_s} = \int \pdata{s\tbar t}{\vect{x}_t}{\vect{x}_s} \pdata{t}{}{\vect{x}_t} \, \rmd \vect{x}_t$. Given a sample $\rdmvect{X}_t \sim \pdata{t}{}{}$, drawing $\rdmvect{X}_s \sim \pdata{s\tbar t}{\rdmvect{X}_t}{}$ yields a sample from the more structured distribution $\pdata{s}{}{}$. 
However, these backward transition kernels are intractable  due to the integration over the posterior $\pdata{0\tbar t}{\vect{x}_t}{}$. Following \cite{ho2020denoising}, the latter is approximated with a Dirac mass at $\denoiser{t}{}{\vect{x}_t}[\theta]$, the neural network approximation of its expectation, \emph{i.e.}, the denoiser $\mathbb{E}[\rdmvect{X}_0 | \rdmvect{X}_t = \vect{x}_t]$ where $\rdmvect{X}_0 \sim \pdata{0}{}{}$ and $\rdmvect{X}_t \sim \fw{t \tbar 0}{\rdmvect{X}_0}{}$. This yields the transition approximation
\begin{equation}
	\label{eq:learned-backward-transition-unconditional}
	\pdata{s\tbar t}{\vect{x}_t}{\vect{x}_s}[\theta] \eqdef \fw{s\tbar 0, t}{\denoiser{t}{}{\vect{x}_t}[\theta], \vect{x}_t}{\vect{x}_s}.
\end{equation}
The parameters $\theta$ of the denoiser are typically learned by minimizing a weighted $\ell_2$ denoising loss over several time steps:
\begin{equation}
\label{eq:unconditional-denoiser-training}
	\min_\theta \textstyle \sum_{t=1}^T w_t \, \mathbb{E} \!\left[ \| \denoiser{t}{}{\rdmvect{X}_t}[\theta] - \rdmvect{X}_0 \|_2^2 \right],
\end{equation}
where $(w_t)_{t=1}^T$ denotes a sequence of non-negative weights.
At inference time, the generative process starts with $\hat{\rdmvect{X}}_T \sim \gauss(0, \Id_d)$ and iteratively samples
$\hat{\rdmvect{X}}_{t-1} \sim \pdata{t-1 \tbar t}{\hat{\rdmvect{X}}_t}{}$ for {$t > 1$}.
Finally, $\hat{\rdmvect{X}}_0 \eqdef \denoiser{1}{}{\hat{\rdmvect{X}}_1}[\theta]$ is used as an approximate sample from $\pdata{0}{}{}$.

\paragraph{Bayesian Inverse Problems.}
In the Bayesian formulation of \emph{inverse problems}, the following observation model is considered:
\begin{equation}
\label{eq:bayesian-formulation-inv-prob}
	\rdmvect{Y} = \operator(\rdmvect{X}_0) + \sigma_{\obs} \rdmvect{W},
	\, \rdmvect{X}_0 \sim \pdata{0}{}{},
	\, \rdmvect{W} \sim \gauss(0, \Id_\dimobs),
\end{equation}
where $\operator$ is a known \emph{degradation operator} (also called \emph{forward model}) acting on the 
{space of the signal to recover (e.g., the pixel space for image inverse problems),}
$\sigma_{\obs}$ is the noise level and $\rdmvect{X}_0$ and $\rdmvect{W}$ are independent.
The probability of observing $\rdmvect{Y}=\obs$ given $\rdmvect{X}_0=\vect{x}_0$ {and a degradation operator $\operator$} is defined as
\begin{equation}
\label{eq:likelihood-measurement}
	\lklh{0}{\vect{x}_0, \operator} \eqdef \gausspdf\!\left(y; \operator(\vect{x}_0), \sigma_{\obs}^2 \Id_\dimobs \right).
\end{equation}
In the Bayesian paradigm, the reconstruction of $\rdmvect{X}_0$ from an observation $\rdmvect{Y}=\obs$ {and the known degradation operator $\operator$} amounts to sampling from the posterior distribution associated with \eqref{eq:bayesian-formulation-inv-prob}:
\begin{equation}
\label{eq:posterior-distribution}
	\posterior{0}{}{\vect{x}_0} \propto \lklh{0}{\vect{x}_0, \operator} \, \pdata{0}{}{\vect{x}_0}.
\end{equation}

\subsection{Diffusion Posterior Sampling} 
The core idea of zero-shot diffusion posterior sampling algorithms \citep{daras2024survey, janati2025bridging, chung2025diffusion} is to approximate the dynamics of a Denoising Diffusion Model (DDM) targeting the posterior distribution \eqref{eq:posterior-distribution} by suitably adapting, at inference time, a DDM trained on the prior $\pdata{0}{}{}$. 
A key property of these methods is that they require no additional training on data generated from the observation model \eqref{eq:bayesian-formulation-inv-prob}, {beyond the training \eqref{eq:unconditional-denoiser-training} of the DDM model targeting the prior $\pdata{0}{}{}$.} 
Constructing a DDM for the posterior distribution $\posterior{0}{}{}$ {for a given known $\obs$ and $\operator$} amounts to approximately sampling along the distributional path $(\posterior{t}{}{})_{t=0}^T$, defined analogously to the {unconditional probability path $(\pdata{t}{}{})_{t=0}^T$}:
\begin{equation}
\label{eq:posterior-sampling-proba-path}
	\posterior{t}{}{\vect{x}_t} := \int \fw{t\tbar 0}{\vect{x}_0}{\vect{x}_t} \posterior{0}{}{\vect{x}_0} \, \rmd \vect{x}_0 \eqsp,
\end{equation}
This is achieved by having a parametric approximation of the posterior denoiser 
$\pE[\rdmvect{X}_0 \mid \rdmvect{X}_t = \vect{x}_t, \rdmvect{Y} = \obs]$, 
where $\rdmvect{X}_0 \sim \pdata{0}{}{}$, $\rdmvect{Y} \sim \lklhwithoutobs{0}{\rdmvect{X}_0, \operator}$, and $\rdmvect{X}_t \sim \fw{t \tbar 0}{\rdmvect{X}_0}{}$. 
It satisfies the identity \citep[Eq. 2.15 and 2.17]{daras2024survey}
\begin{equation}
\label{eq:posterior-denoiser-identity}
\begin{aligned}
\pE[\rdmvect{X}_0 \mid &\rdmvect{X}_t = \vect{x}_t, \rdmvect{Y} = \obs]
= \\
&\pE[\rdmvect{X}_0 \mid \rdmvect{X}_t = \vect{x}_t] + \frac{\sigma_t^2}{\alpha_t}
\nabla_{\vect{x}_t}
\log \lklh{t}{\vect{x}_t, \operator},
\end{aligned}
\end{equation}
with
\begin{equation}
     \lklh{t}{\vect{x}_t, \operator} := \int \lklh{0}{\vect{x}_0, \operator} \, \pdata{0 \tbar t}{\vect{x}_t}{\vect{x}_0} \, \rmd \vect{x}_0.
\end{equation}
Hence, {assuming that we already have pre-trained a neural network $\denoiser{t}{}{\vect{x}_t}[\theta]$ approximating  the unconditional denoiser $\pE[\rdmvect{X}_0 \mid \rdmvect{X}_t = \vect{x}_t] $}, estimating the posterior denoiser $\pE[\rdmvect{X}_0 \mid \rdmvect{X}_t = \vect{x}_t, \rdmvect{Y} = \obs]$ reduces to approximating the intractable log-gradient on the right-hand side of \eqref{eq:posterior-denoiser-identity}. 
A widely used approximation \citep{ho2022video,chung2023diffusion} replaces $\pdata{0 \tbar t}{\vect{x}_t}{}$ in the integral by a Dirac mass centered at the prior denoiser $\denoiser{t}{}{\vect{x}_t}[\theta]$, leading to
\begin{equation}
\label{eq:posterior-gradient-approx}
    \nabla_{\vect{x}_t} \log \lklh{t}{\vect{x}_t, \operator}
    \;\approx\; \nabla_{\vect{x}_t} \log \hlklh{t}{\vect{x}_t, \operator} \eqsp
\end{equation}
with
\begin{equation}
    \hlklh{t}{\vect{x}_t, \operator} \; \eqdef \; \lklh{0}{\denoiser{t}{}{\vect{x}_t}[\theta], \operator}.
\end{equation}
This likelihood-guided correction is the key source of flexibility of zero-shot methods, but also their dominant computational bottleneck, and will be the focus of the amortization strategy introduced in the next section.

\subsection{Supervised Diffusion and Fully Amortized Methods}

To reduce inference-time computation, an alternative to zero-shot posterior sampling is to rely on an upstream training phase, as done in supervised diffusion and fully amortized inference methods. While these approaches enable fast inference, they fundamentally trade flexibility for efficiency, and typically fail to generalize to degradation operators not seen during training.

\paragraph{Supervised Diffusion.}
Supervised diffusion methods \citep{saharia2022image,saharia2022palette} directly learn a neural approximation of the posterior denoiser
$\pE[\rdmvect{X}_0 \mid \rdmvect{X}_t = \vect{x}_t, \rdmvect{Y} = \obs]$
by training a conditional denoiser on paired clean–degraded data.
Inference then proceeds via ancestral sampling using learned, observation-conditioned backward transitions.
Recent extensions condition the denoiser additionally on the degradation operator $\operator$ to handle multiple operators within a single model \citep{elata2025invfussion}.
However, since the likelihood correction is entirely internalized by the network during training, these methods remain tightly coupled to the distribution of operators seen during training and exhibit poor robustness to unseen degradations ({see \Cref{app:lack_robustness_ood_sft} for an experimental illustration).}

\paragraph{Fully Amortized Methods.}
Fully amortized approaches \citep{lee2024diffusion,feng2023score,feng2024variational} rely on variational inference to learn an implicit distribution
$r_\phi(\cdot | \obs, \operator)$ that approximates the target posterior $\posterior{0}{}{}$.
By parameterizing this distribution as the push-forward of a simple base distribution through a neural network, they enable one-step inference at test time.
However, similarly to supervised diffusion, the absence of explicit likelihood guidance at inference makes these methods highly dependent on the training distribution of degradation operators, resulting in limited out-of-distribution generalization.

In contrast to both supervised diffusion and fully amortized inference, our approach retains explicit likelihood-guided posterior sampling at test time, thereby preserving robustness to arbitrary, previously unseen degradation operators.

\section{Method}
\label{sec:method}

We now present our approach to accelerate diffusion posterior sampling while preserving out-of-distribution robustness. Our key insight is to introduce amortization not at the level of the posterior itself, but within the inner variational inference problems solved at each timestep of so-called \emph{variational diffusion posterior sampling} methods. This design leverages upstream training to reduce inference-time computation while retaining explicit likelihood guidance at test 
time, yielding an amortization strategy that, to our knowledge, has 
not been explored in prior work.

\subsection{Variational Diffusion Posterior Sampling}

Variational diffusion posterior sampling \citep{janati2025a,moufad2025variational} denotes a class of zero-shot posterior sampling methods that rely on variational inference to approximate intractable intermediate distributions—such as selected conditional or marginal distributions arising along the diffusion trajectory—thereby enabling approximate backward diffusion toward the target posterior while retaining explicit likelihood guidance.
To illustrate our approach, we focus on a representative instance of this class, MGDM \citep{janati2025a}, but emphasize that the amortization strategy proposed in this work applies more broadly to variational diffusion posterior sampling methods.

\paragraph{Variational Inference in MGDM.}
In \citet{janati2025a}, the key idea is to approximate the marginal distribution $\posterior{t}{}{}$ in \eqref{eq:posterior-sampling-proba-path} by a mixture 
$\approxposterior{t}{}{}$ of midpoint-based approximations $\{ \approxposterior{t}{}{}[s] \}_{s=1}^{t-1}$, i.e.,
$
    \posterior{t}{}{\vect{x}_t} \;\approx\; \approxposterior{t}{}{\vect{x}_t}
    \;\eqdef\; \sum_{s=1}^{t-1} \omega_t^s \, \approxposterior{t}{}{\vect{x}_t}[s],
$
where $(\omega_t^s)_{s=1}^{t-1}$ are non-negative weights summing to $1$, and
\begin{equation}
\label{eq:approximation-mgdm}
	\approxposterior{t}{}{\vect{x}_t}[s] \propto 
	\left( \int \hlklh{s}{\vect{x}_s, \operator} \,
	 \pdata{s \tbar t}{\vect{x}_t}{\vect{x}_s} \, \rmd \vect{x}_s \right) \pdata{t}{}{\vect{x}_t},
\end{equation}
is an approximation of $\posterior{t}{}{\vect{x}_t}$ based on a midpoint timestep $s<t$. In \citet{janati2025a}, approximate sampling from the target posterior distribution 
$\posterior{0}{}{}$ is performed sequentially through the intermediate distributions 
$\approxposterior{T}{}{}, \approxposterior{T-1}{}{}, \ldots, \approxposterior{0}{}{}$. 
At each step $t$, sampling from the mixture $\approxposterior{t}{}{}$ proceeds by first 
drawing a midpoint index $s < t$ according to the categorical distribution with weights $(\omega_t^s)_{s=1}^{t-1}$, 
and then sampling from the corresponding component $\approxposterior{t}{}{}[s]$ in \eqref{eq:approximation-mgdm} 
using a Gibbs sampler with the following data augmentation scheme: 
\begin{equation}
\label{eq:gibbs-data-augmentation}
\begin{split}
\bar{\pi}_{0,s,t}(&\vect{x}_0, \vect{x}_s, \vect{x}_t \mid \obs, \operator) \propto \\
& \pdata{0 \tbar s}{\vect{x}_s}{\vect{x}_0}
\, \hlklh{s}{\vect{x}_s, \operator}
\, \pdata{s \tbar t}{\vect{x}_t}{\vect{x}_s}
\, \pdata{t}{}{\vect{x}_t} .
\end{split}
\end{equation}
By construction, the $\vect{x}_t$-marginal of this joint distribution is exactly 
$\approxposterior{t}{}{}[s]$. 
The Gibbs sampler \citep{gelfand1990sampling} repeats for $R$ steps the following updates: given $(\rdmvect{X}^{r} _0, \rdmvect{X}^{r} _s, \rdmvect{X}^{r} _t)$, 
\begin{enumerate}
	\item Sample $\rdmvect{X}^{r+1} _s$ from 
    \begin{equation}
    \label{eq:conditional-xs-knowing-x0-xt}
    \begin{split}
        \barposterior{s \tbar 0,t}{&\rdmvect{X}^r _0, \rdmvect{X}^r_t}{\vect{x}_s} \\
        &\propto \hlklh{s}{\vect{x}_s, \operator} \, \fw{s \tbar 0, t}{\rdmvect{X}^r _0, \rdmvect{X}^r_t}{\vect{x}_s}.
    \end{split}
    \end{equation}
    In practice, this step is implemented using a Gaussian variational distribution, as we describe below. 
    This approximation is precisely what places the method of \citet{janati2025a} within the class of variational posterior sampling approaches.
	\item Sample $\rdmvect{X}^{r+1} _0$ from $\pdata{0 \tbar s}{\rdmvect{X}^{r+1} _s}{}$ \label{line:mgdm-denoising}
	{using the approximate reverse process of the pre-trained DDM that targets the prior $\pdata{0}{}{}$,  based on the kernels \eqref{eq:learned-backward-transition-unconditional}.}
    \item Sample $\rdmvect{X}^{r+1} _t$ from $\fw{t \tbar s}{\rdmvect{X}^{r+1} _s}{}$, which is a forward Gaussian noising process \eqref{eq:forward-process} in DDMs.
\end{enumerate}

\paragraph{Variational Approximation.}
Given the context variable $\vect{c}:= \context$, 
the variational approximation consists in minimizing
\begin{equation}
\label{eq:variational-app-gaussian}
\begin{split}
    \min_{\vect{\mu}, \vect{\rho}}& \; \mathcal{L}(\vect{\mu}, \vect{\rho}; \vect{c}) \\
	&\eqdef\;
    \KLdiv{\gausspdf(\cdot; \vect{\mu}, \mathrm{diag}(\vect{\rho}))}
	{\barposterior{s \tbar 0,t}{\vect{x}_0, \vect{x}_t}{}},
\end{split}
\end{equation}
where $\KL$ denotes the Kullback–Leibler divergence,
$\barposterior{s \tbar 0,t}{\vect{x}_0, \vect{x}_t}{\vect{x}_s}$ is defined in \eqref{eq:conditional-xs-knowing-x0-xt}, and
$\mathrm{diag}(\vect{\rho})$ is the diagonal matrix with entries given by the vector $\vect{\rho}$. This KL divergence is in practice estimated using the reparameterization trick \citep{kingma2013auto} with a single Monte Carlo sample.
Achieving a sufficiently small value of this objective is critical for obtaining high-quality reconstructions in variational diffusion posterior sampling algorithms \citep{janati2025a}.
However, in practice this minimization is performed at inference time using first-order optimization, which typically requires multiple gradient updates and costly backpropagations through the pre-trained unconditional denoiser $\denoiser{t}{}{}[\theta]$.
The repeated numerical resolution of this variational optimization problem at inference time constitutes the main computational bottleneck of MGDM, and is the target of the amortization strategy introduced next.

\subsection{Amortized Variational Inference}

We propose an \emph{amortized optimization} approach \citep{amos2023tutorial}
that transfers the burden of minimizing the variational objective
\eqref{eq:variational-app-gaussian} from inference to a dedicated upstream training phase.

\paragraph{Inference Model Training.}
The \emph{inference model} is defined as $\vect{c}\mapsto \vect{\lambda}_\varphi(\vect{c}) := (\vect{\mu}_\varphi(\vect{c}), \vect{\rho}_\varphi(\vect{c}))$, with parameters~$\varphi$, to predict a good initialization to address the variational approximation problem \eqref{eq:variational-app-gaussian}, given a new context $\vect{c}$. The model is implemented using a single forward pass through a neural network and trained by minimizing the objective-based loss \citep{amos2023tutorial}:
\begin{equation}
\label{eq:inference-model-training}
    \min_\varphi \mathbb{E}_{\vect{c}} \left[ \mathcal{L}( \vect{\mu}_\varphi(\vect{c}), \vect{\rho}_\varphi(\vect{c}); \vect{c}) \right],
\end{equation}
where $\mathcal{L}$ is the variational objective from \eqref{eq:variational-app-gaussian}. 
The training loss \eqref{eq:inference-model-training} is minimized by stochastic gradient descent. 
The context $\vect{c}:= \context$ in the expectation of \eqref{eq:inference-model-training} is a random variable drawn from a distribution designed to match those encountered at inference, \emph{i.e.}, during posterior sampling with MGDM. We find the following sampling procedure for $\vect{c}$ to be effective in practice: 
\begin{gather}
(\vect{x}, \obs, \operator) \sim
\pdata{\textrm{data}}{}{\vect{x}}\,
\lklh{0}{x, \operator}\,
\pdata{\textrm{op}}{}{\operator}, \label{eq:sampling-x-y-op}\\
\begin{aligned}
t &\sim \mathrm{Unif}(\mathcal{T}), &
s \mid t &\sim \mathrm{Categorical}\!\left( (\omega_t^{s'})_{s'=1}^{\,t-1} \right), \\
\vect{x}_t \mid \vect{x} &\sim \fw{t \tbar 0}{\vect{x}}{}, &
\vect{x}_0 &:= \denoiser{t}{}{\vect{x}_t}[\theta].
\end{aligned}
\end{gather}
where $\pdata{\textrm{data}}{}{\vect{x}}$ denotes the distribution of clean images, $\pdata{\textrm{op}}{}{\operator}$ denotes the distribution of degradation operators modeled during training, and $\mathcal{T} \subseteq \{ 1, \ldots, T \}$ denotes the subset of timesteps on which the inference model is applied during test-time inference. 

\paragraph{Inference Model Parameterization.}
The inference model aims to predict a near minimizer of problem \eqref{eq:variational-app-gaussian}, so
it is natural to start from the initialization used in MGDM \citep{janati2025a},
where the optimization is initialized with the
statistics of the Gaussian bridge transition $\fw{s\tbar 0,t}{\vect{x}_0,\vect{x}_t}{\vect{x}_s} \propto
 \fw{s \tbar 0}{\vect{x}_0}{\vect{x}_s}\fw{t \tbar s}{\vect{x}_s}{\vect{x}_t}$ and then refined by gradient descent.
To mimic this procedure in a single forward pass,
we parameterize the network so that it learns the \emph{residual}
between these prior statistics and the posterior parameters that
approximately minimize \eqref{eq:variational-app-gaussian}.

\paragraph{Inference Model Architecture.}
Concerning the architecture of this neural network,
note that minimizing the variational objective
\eqref{eq:variational-app-gaussian}
amounts to approximately sampling from the conditional law
$\barposterior{s \tbar 0,t}{\vect{x}_0,\vect{x}_t}{\vect{x}_s}
 \propto
 \hlklh{s}{\vect{x}_s,\operator}\,
 \fw{s \tbar 0,t}{\vect{x}_0,\vect{x}_t}{\vect{x}_s}$ defined in
\eqref{eq:conditional-xs-knowing-x0-xt}, given a context $\vect{c}= \context$.
In other words, the task is to \emph{denoise} from $\vect{x}_t$ down to an
intermediate state $\vect{x}_s$ with $s<t$,
while conditioning on $\vect{x}_0$ and, crucially,
on the observation $\obs$ and the operator $\operator$.
This conditioning naturally suggests drawing inspiration from the
conditional denoisers used in supervised diffusion models. To efficiently account for multiple degradations during training, we recommend adopting the InvFussion \citep{elata2025invfussion} architecture,
which is expressly designed to handle external inputs such as $\obs$ and $\operator$.
Therefore, the neural-network architecture of our inference model can be derived from the backbone of any existing conditional denoisers, with the following minor modifications:
(i) concatenate $\vect{x}_0$, $\vect{x}_t$ along the channel dimension;
(ii) early fusion of the embedding of the timesteps $t$ and $s$;
(iii) double the number of output channels so the network jointly predicts variational parameters that approximately minimize \eqref{eq:variational-app-gaussian}.\footnote{These minor changes are compatible with all existing standard conditional denoiser architectures, e.g., UNet \citep{ho2020denoising}, DiT \citep{peebles2023scalable}, HDiT \citep{crowson2024scalable}.} 
Further details are deferred to \Cref{app:details-inference-model}. 

\subsection{Sampling with Amortized Initialization}

We now summarize the new sampling algorithm, {called \ours (Likelihood-guided Amortized Variational Posterior Sampling)}, as it uniquely combines amortization and explicit likelihood-guidance.

\paragraph{Amortized Warm-start with Likelihood Guidance.}
At each diffusion timestep, we use the trained inference model to warm-start the optimization of \eqref{eq:variational-app-gaussian}. 
Given a context $\vect{c}:= (\vect{x}_0, \vect{x}_t, s, t, \obs, \operator)$, the model predicts $(\vect{\mu}_\varphi(\vect{c}), \vect{\rho}_\varphi(\vect{c}))$ in a single forward pass, which is used to initialize a first-order optimizer. 
This initialization is tailored to the inverse problem at hand by explicitly accounting for the observation $\obs$ and degradation operator $\operator$, and allows the variational objective \eqref{eq:variational-app-gaussian} to be minimized using only a small number of likelihood-guided gradient steps, without compromising final reconstruction quality.

\begin{figure*}[t]
    \centering
    \begin{subfigure}[t]{0.33\linewidth}
        \centering
        \includegraphics[width=\linewidth]{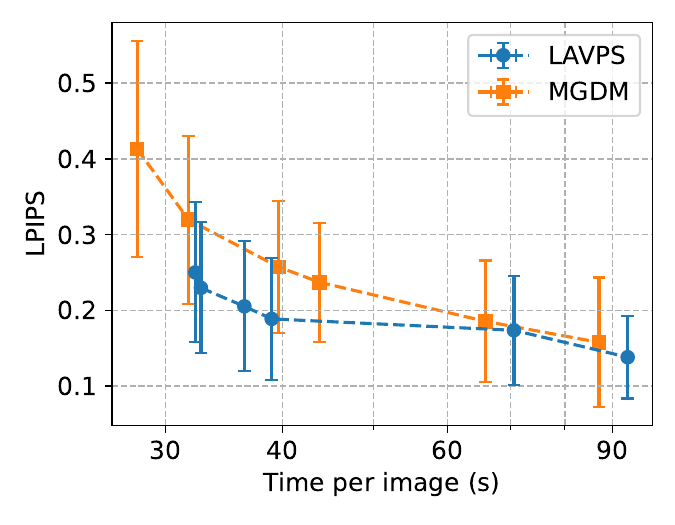}
        \caption{Motion deblurring}
        \label{fig:pareto_in_mb21_lpips}
    \end{subfigure}
    \hfill
    \begin{subfigure}[t]{0.33\linewidth}
        \centering
        \includegraphics[width=\linewidth]{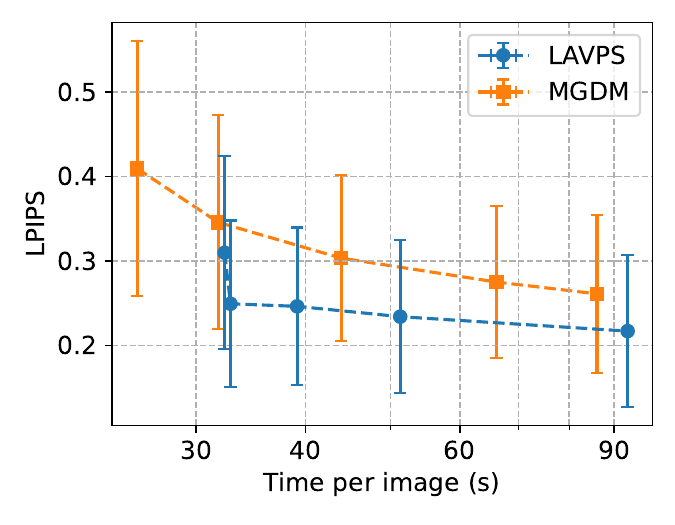}
        \caption{Super-resolution $\times 4$}
        \label{fig:pareto_in_sr4_lpips}
    \end{subfigure}
        \hfill
    \begin{subfigure}[t]{0.33\linewidth}
        \centering
        \includegraphics[width=\linewidth]{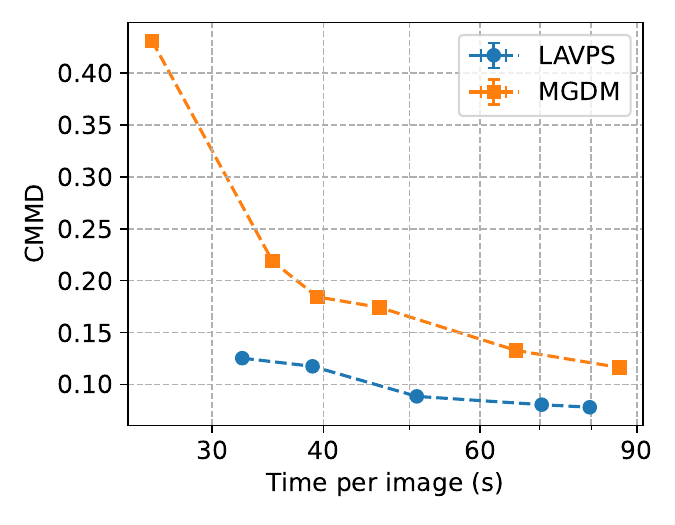}
        \caption{Inpainting}
        \label{fig:pareto_in_ip2_cmmd}
    \end{subfigure}

    \caption{Efficiency-quality trade-off for \ours vs. MGDM, on in-distribution degradation operators. Markers represent Pareto-optimal configurations for each method across various restoration tasks on the \texttt{Imagenet} dataset.}
    \label{fig:pareto_in}
\end{figure*}

\paragraph{Safeguarded Initialization for Robustness.}
To preserve robustness when the degradation operator at test time is out-of-distribution, we introduce a simple safeguard at initialization. 
Specifically, we compare the variational objective evaluated at the warm-start initialization, denoted by $\mathcal{L}_{\mathrm{ws}}$, with its value under the standard zero-shot initialization used in MGDM \citep{janati2025a}, i.e., an initialization derived from the unconditional prior bridge and independent of the observation $\obs$ and degradation operator $\operator$, denoted by $\mathcal{L}_{\mathrm{zs}}$.
If $\mathcal{L}_{\mathrm{ws}} > \mathcal{L}_{\mathrm{zs}}$, indicating that the inference model may be unreliable, we discard the warm-start and revert to the zero-shot initialization. In practice, our experiments show that this fallback mechanism is sufficient to ensure robustness to arbitrary degradation operators, including those unseen during training, and guarantees performance no worse than that of MGDM in terms of reconstruction quality.

\section{Related Work}
Two recent works \citep{elata2025invfussion,mbakam2025learning} share with us the idea of training a component that can be reused at inference to remain flexible to different likelihoods (equivalently, different degradation operators), but they differ in how the likelihood is ultimately exploited: they predict a likelihood-conditioned denoised sample \emph{directly through a network} that takes the likelihood parameters as input.  \citet{elata2025invfussion} adapt deep unrolling \citep{gregor2010learning} by training a conditional diffusion model with an operator-aware block that applies the degradation operator $\operator$ and its pseudo-inverse directly to the activations, enabling the network to predict a likelihood-conditioned denoised sample.
\citet{mbakam2025learning} unfold the LATINO Langevin sampler \citep{spagnoletti2025latino} into a sequence of learnable steps and train it with a consistency trajectory loss so that the network likewise predicts a likelihood-conditioned denoised sample.
Because these methods rely on the network to internalize the likelihood correction, they may require large paired datasets and  generalize poorly to degradation operators not covered during training.

In contrast, our approach keeps likelihood-guided optimization explicit: the inference model provides an approximate initialization, while the final correction is obtained by directly minimizing the variational objective \eqref{eq:variational-app-gaussian} under the true test-time likelihood. This combination of amortized initialization and explicit likelihood guidance enables both fast inference and improved robustness to previously unseen operators, as demonstrated in the next section.

\section{Experiments}
\label{sec:experiments}

Our protocol follows \citet{elata2025invfussion}.
The unconditional denoiser $\denoiser{t}{}{}[\theta]$ is an HDiT \citep{crowson2024scalable} denoiser. We validate our method both on facial images using the \texttt{FFHQ} dataset \citep{karras2019style} and natural images using the \texttt{ImageNet} \citep{deng2009imagenet}, at the resolution of $256 \times 256$.
Methods are evaluated on diverse image restoration tasks: super-resolution, inpainting, and motion deblurring. The noise level in  \eqref{eq:bayesian-formulation-inv-prob} is fixed at $\sigma_{\obs} = 0.05$. The evaluation is performed on $300$ test images, unseen during training or hyperparameter tuning. The reconstruction quality is measured with per-image metrics comparing the similarity between the reconstructed image and a reference one, such as LPIPS \citep{zhang2018unreasonable}, SSIM \citep{wang2004image} and PSNR. We report the average of these metrics with their standard deviations over all test images.
To measure the photorealism of the reconstructed images, we also report the CMMD\footnote{With only 300 images, we found FID \citep{heusel2017gans} to be unstable and poorly aligned with perceptual quality. We therefore rely on CMMD, which is more sample-efficient and better suited to our experiments.} \citep{jayasumana2024rethinking}, {which is particularly appropriate for inpainting tasks, where multiple valid reconstructions exist and per-instance fidelity metrics are inherently ambiguous.} All details about the experimental protocol are deferred to \Cref{app:details-experimental-protocol}.

\subsection{Acceleration for In-distribution Operators}
\label{subsec:acceleration_id}

{We first show that \ours improves the trade-off between reconstruction quality and inference speed compared to zero-shot posterior sampling methods, thanks to the proposed amortization strategy. To isolate this effect, we first consider a simplified setting in which amortization and evaluation are performed on the \emph{same} degradation operator. The more general scenario—where amortization is carried out over a \emph{family} of operators and robustness to out-of-distribution degradations is assessed—is deferred to \Cref{subsec:robustesse_ood}.}

{As shown by \citet{janati2025a}, MGDM is a strong baseline that achieves a good trade-off between quality and speed among zero-shot posterior sampling methods (cf. \Cref{app:ffhq256_dps} for details). Therefore, our goal here is to show that \ours achieves a better trade-off than MGDM.}

\begin{table}[t]
\centering
\caption{Inference speedup of \ours vs. MGDM for matched reconstruction quality levels on \texttt{FFHQ} and \texttt{ImageNet} ($256 \times 256$). We report the sampling time required to reach a specific quality target (LPIPS, or CMMD denoted by $\dagger$) measured on a single NVIDIA V100.}
\label{tab:speedup_results_compact}
\resizebox{\columnwidth}{!}{%
\begin{tabular}{llccc}
\toprule
\textbf{Task} & \textbf{Dataset} &
\begin{tabular}[c]{@{}c@{}}\textbf{Target}\\\textbf{Quality $\downarrow$}\end{tabular} &
\begin{tabular}[c]{@{}c@{}}\textbf{Inference time}\\\textbf{of MGDM}\end{tabular} &
\begin{tabular}[c]{@{}c@{}}\textbf{Speedup}\\\textbf{with \ours}\end{tabular} \\ 
\midrule
\multirow{4}{*}{SR $\times 4$} & \multirow{2}{*}{\texttt{FFHQ}} & 0.102 & 29.3s & $\mathbf{1.73\times}$ \\
 &  & 0.096 & 44.5s & $\mathbf{1.41\times}$ \\ \cmidrule{2-5}
 & \multirow{2}{*}{\texttt{ImageNet}} & 0.275 & 66.1s & $\mathbf{2.01\times}$ \\
 &  & 0.261 & 86.1s & $\mathbf{2.15\times}$ \\ \midrule
\multirow{4}{*}{Deblurring} & \multirow{2}{*}{\texttt{FFHQ}} & 0.106 & 19.2s & $\mathbf{1.27\times}$ \\
 &  & 0.062 & 48.3s & $\mathbf{1.25\times}$ \\ \cmidrule{2-5}
 & \multirow{2}{*}{\texttt{ImageNet}} & 0.257 & 39.6s & $\mathbf{1.20\times}$ \\
 &  & 0.237 & 43.8s & $\mathbf{1.34\times}$ \\ \midrule
\multirow{3}{*}{Inpainting} & \multirow{3}{*}{\texttt{ImageNet}} & 0.174$^\dagger$ & 46.2s & $\mathbf{1.43\times}$ \\
 &  & 0.133$^\dagger$ & 65.9s & $\mathbf{2.03\times}$ \\
 &  & 0.116$^\dagger$ & 86.0s & $\mathbf{2.21\times}$ \\ \bottomrule
\end{tabular}%
}
\end{table}

\paragraph{Pareto Frontier Analysis.} 
\Cref{fig:pareto_in} shows that LAVPS consistently shifts the Pareto frontier of MGDM downward for the \texttt{ImageNet} dataset, on super-resolution $\times 4$, motion deblurring with a kernel size $21 \times 21$, and the inpainting task where the mask is a center box of size $128 \times 128$.
{\Cref{app:additional_acceleration_id_pareto_ffhq} shows the same conclusion for the \texttt{FFHQ} dataset.}
The gains of \ours are most pronounced in the low-to-medium computational budget regime\footnote{As the inference budget increases, the performance of both methods converges, which is expected: given sufficient compute, the iterative optimization in MGDM can eventually compensate for its coarse initialization and reach the same variational optimum as \ours.}, where amortized initialization enables high-fidelity reconstructions at substantially reduced sampling cost. This is illustrated in \Cref{fig:figure-intro}, where \ours recovers crisp high-frequency details—such as the bird's feathers and the shark's teeth—whereas the zero-shot baseline produces significantly blurred or noisy reconstructions that require considerably more computational budget to reach a similar level of fidelity. See \Cref{app:additional_acceleration_id_gradient_steps} for an ablation study analyzing the role of amortized initialization in reducing the number of inner-loop gradient steps required to achieve high-quality reconstructions.

\paragraph{Quantifying Speedups.} 
In \Cref{tab:speedup_results_compact}, we quantify the acceleration of \ours at inference compared to MGDM, when considering different reconstruction quality target along the Pareto frontier of MGDM.
Specifically, we identify the fastest \ours configuration that matches the quality of representative points from the baseline, validating performance parity through a statistical non-inferiority test. This procedure ensures that the reported speedups represent true efficiency gains without sacrificing reconstruction quality. 
Our results demonstrate systematic speedups across all tasks:
\begin{itemize}[leftmargin=*, nolistsep, topsep=0pt]
    \item On \texttt{FFHQ}, \ours achieves targeted qualities $1.25\times$ to $1.73\times$ faster for motion deblurring and super-resolution.
    \item On \texttt{ImageNet}, evaluations show higher speedups, up to $2.03\times$ for inpainting and up to $2.15\times$ for super-resolution.
\end{itemize}
{These inference-time gains come at the cost of an additional amortization training phase, whose computational overhead is quantified in \Cref{app:ffhq256_computational_cost}. In practice, this cost is modest—only a few hours on a single V100 GPU—making the resulting inference speedups highly favorable.}

\begin{table}[t]
\centering
\caption{Motion deblurring on \texttt{ImageNet}: Improvement on in-distribution while keeping out-of-distribution robustness.}
\label{tab:placeholder_label}
\small
\resizebox{\columnwidth}{!}{%
\begin{tabular}{l|cc|cc}
\toprule
 & \multicolumn{2}{c|}{\textbf{In-distribution operators}} & \multicolumn{2}{c}{\textbf{Out-of-distribution operators}} \\
  & \multicolumn{2}{c|}{(kernel size $21 \times 21$)} & \multicolumn{2}{c}{(kernel size $35 \times 35$)} \\
Method & Time (s) & LPIPS $\downarrow$ & Time (s) & LPIPS $\downarrow$ \\
\midrule
\multicolumn{5}{l}{\textbf{Inference time $\leq$ 35.1 s}} \\
MGDM & 35.1 & 0.250 & 35.1 & \textbf{0.278} \\
LAVPS & \textbf{33.2} & \textbf{0.236} & \textbf{34.1} & 0.279 \\
\midrule
\multicolumn{5}{l}{\textbf{Inference time $\leq$ 47.3 s}} \\
MGDM & 45.9 & 0.210 & 39.5 & 0.256 \\
LAVPS & \textbf{40.3} & \textbf{0.188} & \textbf{37.5} & \textbf{0.238} \\
\bottomrule
\end{tabular}
}
\end{table}

\begin{table}[t]
\centering
\caption{Inpainting on \texttt{ImageNet}: Improvement on in-distribution while keeping out-of-distribution robustness.}
\label{tab:placeholder_label_2}
\small
\resizebox{\columnwidth}{!}{%
\begin{tabular}{l|cc|cc}
\toprule
 & \multicolumn{2}{c|}{\textbf{In-distribution operators}} & \multicolumn{2}{c}{\textbf{Out-of-distribution operators}} \\
  & \multicolumn{2}{c|}{(missing rectangles)} & \multicolumn{2}{c}{(missing pixels)} \\
Method & Time (s) & CMMD $\downarrow$ & Time (s) & CMMD $\downarrow$ \\
\midrule
\multicolumn{5}{l}{\textbf{Inference time $\leq$ 34.9 s}} \\
MGDM & 34.8 & 0.199 & 34.9 & 0.336 \\
LAVPS & \textbf{32.3} & \textbf{0.124} & \textbf{32.7} & \textbf{0.335} \\
\midrule
\multicolumn{5}{l}{\textbf{Inference time $\leq$ 102.6 s}} \\
MGDM & 66.3 & 0.113 & 102.6 & \textbf{0.130} \\
LAVPS & \textbf{43.8} & \textbf{0.105} & \textbf{93.2} & 0.131 \\
\bottomrule
\end{tabular}
}
\end{table}

\begin{figure*}[t]
    \centering
    \begin{subfigure}[t]{0.48\textwidth}
        \centering
        \includegraphics[width=\linewidth]{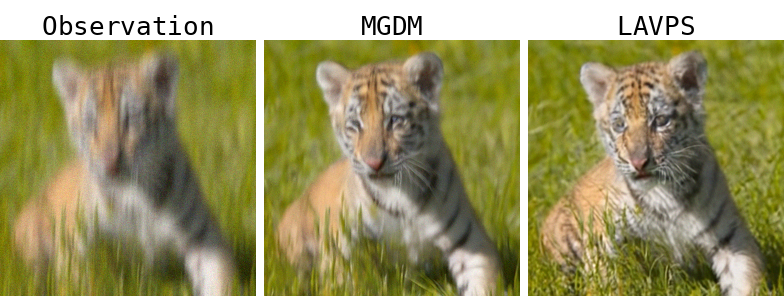}
        \caption{\textbf{In-distribution (kernel size $21 \times 21$)}. MGDM is blurry on the background and the animal.}
    \end{subfigure}
    \hfill
    \begin{subfigure}[t]{0.48\textwidth}
        \centering
        \includegraphics[width=\linewidth]{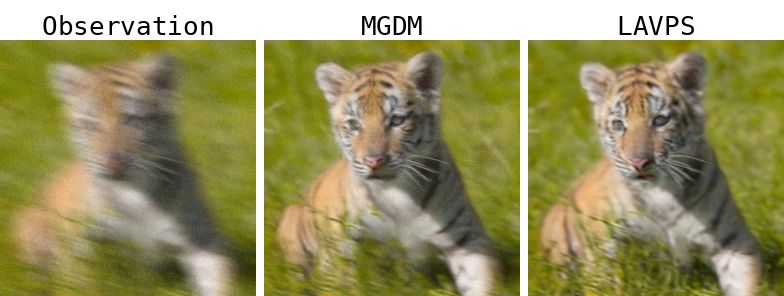}
        \caption{\textbf{Out-of-distribution (kernel size $35 \times 35$)}. \ours does not degrade compared to MGDM.}
    \end{subfigure}
    \vspace{-0.5em}
    \caption{Motion deblurring on \texttt{ImageNet}. 
    The inference time is constrained to be lower than $47.3$s, cf.~\Cref{fig:mb_qualitative_47p3_appendix} for more images.}
    \label{fig:mb_qualitative_47p3}
\end{figure*}

\begin{figure*}[t]
    \vspace{-0.5em}
    \centering
    \begin{subfigure}[t]{0.48\textwidth}
        \centering
        \includegraphics[width=\linewidth]{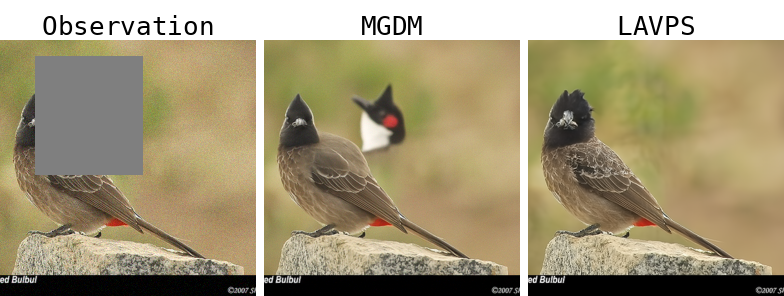}
        \caption{\textbf{In-distribution (missing rectangle)}. MGDM produces hallucinated content, introducing an additional head.}
    \end{subfigure}
    \hfill
    \begin{subfigure}[t]{0.48\textwidth}
        \centering
        \includegraphics[width=\linewidth]{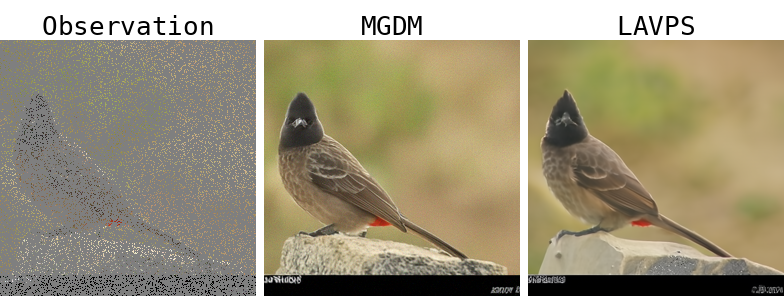}
        \caption{\textbf{Out-of-distribution (missing pixel)}. \ours does not degrade compared to MGDM.}
    \end{subfigure}
    \vspace{-0.5em}
    \caption{
    Inpainting on \texttt{ImageNet}. 
    The inference time is constrained to be lower than $34.9$s, cf.~\Cref{fig:inpainting_qualitative_34p9_appendix} for more images.
    }
    \label{fig:inpainting_qualitative_34p9}
\end{figure*}

\subsection{Robustness to Out-of-distribution Operators}
\label{subsec:robustesse_ood}

We now show that the improvement of the trade-off between quality and inference speed with \ours does not come at the expense of losing the robustness to out-of-distribution degradation operators. 

Our protocol consists in training an inference model by amortizing over a \emph{family} of degradation operators, extending the setting of \Cref{subsec:acceleration_id}, where amortization was performed for a single fixed operator.
Formally, during training, operators are sampled from a fixed distribution $\pdata{\textrm{op}}{}{\operator}$, which we refer to as the \emph{in-distribution} setting. Degradation operators encountered at test time that fall outside this distribution are referred to as \emph{out-of-distribution}.
At inference time, we are given a reconstruction task for which the degradation operator $\mathcal{A}$ is known, consistent with standard assumptions in zero-shot diffusion posterior sampling. However, we do not assume access to information about whether $\mathcal{A}$ is in-distribution or not, as is standard when evaluating robustness to distributional shift.
Evaluation of \ours and MGDM are then performed on diverse degradation operators (in-distribution or out-of-distribution).

\paragraph{Tasks.}
We consider two reconstruction tasks on \texttt{ImageNet}.
The first task is motion deblurring, where degradation operators are  convolutions with kernels generated from random motion trajectories whose length and curvature are controlled by an intensity parameter.\footnote{Kernels are rendered as piecewise-smooth paths, smoothed, downsampled, and normalized to form valid point spread functions.} In-distribution kernels have size $21 \times 21$ and intensity $0.9$, while out-of-distribution kernels have size $35 \times 35$ and intensity $0.3$.
The second task is inpainting, where in-distribution masks consist of rectangular regions at random locations, with width and height uniformly sampled between $0.4$ and $0.6$ times the image resolution, and out-of-distribution masks are generated by independently removing individual pixels according to a Bernoulli distribution with parameter $p \in [0.8, 0.85]$.

\paragraph{Results.}
When amortizing over a family of degradation operators, \ours continues to achieve a better trade-off between reconstruction quality and inference time than MGDM for in-distribution operators.
This is shown by the Pareto fronts reported in \Cref{fig:imagenet256_motiondeblurring_ood} (\Cref{app:additional_exp_section_5p2}) and is consistent with the fixed-operator setting of \Cref{subsec:acceleration_id}, showing that the efficiency gains of amortization are preserved when training over multiple operators rather than a single one. 

\Cref{tab:placeholder_label,tab:placeholder_label_2} provide quantitative comparisons under fixed inference-time budgets for both in-distribution and out-of-distribution operators, with an illustration in \Cref{fig:mb_qualitative_47p3,fig:inpainting_qualitative_34p9}.
The conclusion is that \ours consistently improves reconstruction quality over MGDM on in-distribution operators for comparable or lower inference times, and the performance on out-of-distribution kernels remains comparable to MGDM across budgets, with no systematic degradation. 
Overall, these results indicate that amortizing the inner optimization in variational diffusion posterior sampling improves efficiency on in-distribution operators without compromising the robustness to previously unseen degradation operators that characterizes zero-shot posterior sampling.

\section{Conclusion}
We introduced a novel amortization strategy that accelerates diffusion posterior sampling methods while preserving their ability to handle arbitrary degradation operators at test time. Our approach amortizes the inner variational inference problems arising in variational diffusion posterior sampling, enabling a principled combination of amortization with explicit likelihood guidance. 
Overall, this work demonstrates that amortization and likelihood-guided sampling can be effectively reconciled, paving the way for diffusion-based inverse problem solvers that are both fast at inference and flexible across a wide range of degradation operators.

\section*{Impact Statement}
This paper presents work whose goal is to advance the field of machine learning. There are many potential societal consequences of our work, none of which we feel must be specifically highlighted here.

\bibliography{bibliography}
\bibliographystyle{icml2026}

\newpage
\appendix
\onecolumn

\section{Additional Details on the Method (\Cref{sec:method})}
\label{app:details-inference-model}

We recall that the context variable $\vect{c}:= \context$ in \eqref{eq:variational-app-gaussian} is a tuple containing images $\vect{x}_0$ and $\vect{x}_t$ at time $0$ and $t$, the two timesteps $t$ and $s$, the observation $\obs$ and the degradation operator $\operator$.

\paragraph{Parameterization of the Inference Model.} We parameterize the inference model using a neural network $\vect{c}\mapsto \vect{f}_\varphi(\vect{c}) := (\vect{f}_\varphi^\text{mean}(\vect{c}), \vect{f}_\varphi^\text{var}(\vect{c}))$ that predicts the residual between the mean, covariance statistics $(\vect{\mu}_{s \tbar 0, t}(\vect{x}_0, \vect{x}_t), \sigma^2_{s \tbar 0, t} \Id_d)$ of the prior Gaussian transition $\fw{s\tbar 0, t}{\vect{x}_0, \vect{x}_t}{\vect{x}_s} \propto  \fw{s\tbar 0}{\vect{x}_0}{\vect{x}_s} \fw{t\tbar s}{\vect{x}_s}{\vect{x}_t}$, and the target posterior statistics $(\vect{\mu}_\varphi(\vect{c}), \vect{\rho}_\varphi(\vect{c}))$ that warm-start the variational approximation problem \eqref{eq:variational-app-gaussian}:
\begin{equation}
    \begin{split}
        \vect{\mu}_\varphi(\vect{c}) &:= \vect{\mu}_{s \tbar 0, t}(\vect{x}_0, \vect{x}_t) + \vect{f}_\varphi^\text{mean}(\vect{c}), \\
        \vect{\rho}_\varphi(\vect{c}) &:= \sigma^2_{s \tbar 0, t} \oned{d} + \vect{f}^{\text{var}}_\varphi(\vect{c}),
    \end{split}
\end{equation}
where $\oned{d}$ denotes the vector full of ones, 
\begin{equation}
\label{eq:prior-mean-init-zs}
    \vect{\mu}_{s \tbar 0, t}(\vect{x}_0, \vect{x}_t) := \gamma_{t \tbar s} \alpha_{s \tbar 0} \vect{x}_0 + (1 - \gamma_{t \tbar s}) \alpha_{t \tbar s}^{-1} \vect{x}_t
\end{equation}
with $\gamma_{t \tbar s} := \sigma^2_{t \tbar s} / \sigma^2_{t \tbar 0}$,
and 
\begin{equation}
\label{eq:prior-var-init-zs}
    \sigma^2_{s \tbar 0, t} := \sigma^2_{t \tbar s} \sigma^2_{s \tbar 0} / \sigma^2_{t \tbar 0},
\end{equation}
because by definition, $\fw{s\tbar 0, t}{\vect{x}_0, \vect{x}_t}{\vect{x}_s} \propto  \fw{s\tbar 0}{\vect{x}_0}{\vect{x}_s} \fw{t\tbar s}{\vect{x}_s}{\vect{x}_t}$, and $\fw{s\tbar 0}{\vect{x}_0}{\vect{x}_s} := \gausspdf\left(\vect{x}_s; \alpha_{s\tbar 0} \vect{x}_0, \sigma^2 _{s\tbar 0} \Id_d \right)$ and $\fw{t\tbar s}{\vect{x}_s}{\vect{x}_t} := \gausspdf\left(\vect{x}_t; \alpha_{t\tbar s} \vect{x}_s, \sigma^2 _{t\tbar s} \Id_d \right)$, by definition \eqref{eq:forward-process}.

The choice of this parameterization is motivated by the fact the MGDM \citep{janati2025a} proposes to initialize the variational approximation problem \eqref{eq:variational-app-gaussian} at inference time using the prior statistics $\vect{\mu}_{s \tbar 0, t}(\vect{x}_0, \vect{x}_t)$ and $\sigma^2_{s \tbar 0, t}$, which are updated using gradient descent updates rules.  

\paragraph{Architecture of the Inference Model.}
The architecture of the neural network  model $\vect{c}\mapsto \vect{f}_\varphi(\vect{c}) := (\vect{f}_\varphi^\text{mean}(\vect{c}), \vect{f}_\varphi^\text{var}(\vect{c}))$ is inspired by the standard design of conditional denoisers used in supervised diffusion. Such denoisers take as inputs the noised image $\vect{x}_t$, the timestep $t$, the observation $\obs$ and the degradation operator $\operator$, and output a prediction of the denoised image \citep{saharia2022image,saharia2022palette,elata2025invfussion}. In our setting, the neural network model involved in the parameterization of the inference model requires two additional inputs: the midpoint timestep $s < t$ and the condition on the reconstructed image $\vect{x}_0$ at timestep $0$. Its outputs are the residual $\vect{f}_\varphi^\text{mean}(\vect{c})$ of the approximate Gaussian distribution and the residual diagonal entries $\vect{f}_\varphi^\text{var}(\vect{c})$ of its covariance, both having the same dimensionality as an image. These minor changes are compatible with all existing standard conditional denoiser architectures, e.g., UNet, DiT, HDiT, etc.

There are many possible modifications of the standard conditional denoiser to accommodate these changes in inputs and outputs. We found the following design to be effective in practice: (i) concatenate $\vect{x}_0$ and $\vect{x}_t$ channel-wise, thereby doubling the input channels; (ii) embed the midpoint timestep $s$ analogously to the embedding of $t$, but independently (e.g., using a separate linear layer), and add both timestep embeddings at the beginning of the architecture (early fusion); (iii) double the number of output channels to jointly predict the Gaussian mean and diagonal covariance entries. 

In all our experiments, the backbone of the neural network for the inference model is an HDiT \citep{crowson2024scalable}. Specifically, in \Cref{subsec:acceleration_id}, we adopt the design of Palette \citep{saharia2022palette} using the HDiT architecture. In \Cref{subsec:robustesse_ood}, where specified, we follow the InvFussion approach \citep{elata2025invfussion} by incorporating so-called feature degradation layers into the HDiT backbone. These layers allow the architecture to adapt to different degradation operators during training and inference.

\paragraph{Training Algorithm and Design of the Context Distribution.} The training algorithm of the inference model is summarized in \Cref{alg:training}.
We now comment the choice of the context distribution involved in the training objective function \eqref{eq:inference-model-training} for amortization. The sampling of $(\vect{x}, \obs, \operator)$ creates a tuple of a clear image, its degraded observation along with the corresponding degradation operator, following the model \eqref{eq:bayesian-formulation-inv-prob}. The sampling of the timesteps $(t, s)$ with $s < t$ aims to match those encountered during posterior sampling, cf.~\Cref{sec:background}. Finally, the choice of the distribution for $(\vect{x}_t, \vect{x}_0)$ aims to capture the fact that, during posterior sampling with MGDM, $\vect{x}_t$ is a noised sample, and $\vect{x}_0$ is a denoised version of $\vect{x}_t$: for this reason, we choose to set $\vect{x}_0 := \denoiser{t}{}{\vect{x}_t}[\theta]$ rather $\vect{x}_0 := \vect{x}$, where $\vect{x}$ is sampled directly from $\pdata{\text{data}}{}{}$, the distribution of clean images.

\begin{algorithm}[h]
\caption{Training fully-amortized inference model}
\label{alg:training}
\begin{algorithmic}[1]
\REQUIRE Dataset of clean images $\mathcal{D}$, distribution of operators $\pdata{\text{op}}{}{}$, subset $\mathcal{T} \subseteq \{ 1, \ldots, T \}$ of timesteps, learning rate $\alpha > 0$
\ENSURE Trained inference model $\vect{\lambda}_\varphi$
\STATE $\varphi \gets$ initialization from scratch 
\FOR{each iteration}
    \STATE Sample clean images $\vect{x} \sim \mathcal{D}$
    \STATE Sample degradation operator $\operator \sim \pdata{\text{op}}{}{}$ 
    \STATE Sample observation $\obs \sim \lklh{0}{\vect{x}, \operator}$
    \STATE Sample timestep $t \sim \mathrm{Unif}(\mathcal{T})$
    \STATE Sample midpoint timestep $s \sim \mathrm{Categorical}\!\left( (\omega_t^{s'})_{s'=1}^{\,t-1} \right)$,
    \STATE Sample $\vect{x}_t \sim \fw{t \tbar 0}{\vect{x}}{}$ \hfill \COMMENT{Forward noising}
    \STATE $\vect{x}_0 \gets \denoiser{t}{}{\vect{x}_t}[\theta]$ \hfill \COMMENT{Pre-trained unconditional denoiser}
    \STATE $\vect{c}\gets \context$
    \STATE Compute gradient $g \gets \nabla_\varphi \mathcal{L}(\vect{\mu}_\varphi(\vect{c}), \vect{\rho}_\varphi(\vect{c}); \vect{c})$
    \STATE Update parameters $\varphi \leftarrow \varphi - \alpha g$
\ENDFOR
\RETURN Inference model $\vect{\lambda}_\varphi := (\vect{\mu}_\varphi(\vect{c}), \vect{\rho}_\varphi(\vect{c}))$
\end{algorithmic}
\end{algorithm}

\paragraph{Inference Algorithm.}
\Cref{alg:inference} summarizes how the trained inference model is used during our method called \ours. 
The three conditional distributions involved in the Gibbs sampling of \Cref{sec:background}, as described below \eqref{eq:gibbs-data-augmentation}, are implemented in \Cref{alg:inference}.

\begin{algorithm}[h]
\caption{\ourmethod}
\label{alg:inference}
\begin{algorithmic}[1]
\REQUIRE Likelihood $\lklh{0}{\cdot, \operator}$ defined from \eqref{eq:likelihood-measurement} with observation $\obs$ and operator $\operator$, unconditional denoiser $(\vect{x}, t) \mapsto \denoiser{t}{}{\vect{x}}[\theta]$ pre-trained via \eqref{eq:unconditional-denoiser-training}, inference model $\vect{c}\mapsto \vect{\lambda}_\varphi(\vect{c})$ trained via \eqref{eq:inference-model-training}, number of timesteps K, timesteps $(t_k)_{k=0}^K$ with $t_K=T$ and $t_0=0$, Gibbs repetition $R$, number of denoising steps $M$, learning rates $\eta$, gradient steps $(G_k)_{k=1}^K$, probabilities $(\omega_k^\ell)_{k \in \{2, \ldots, K-1\}, \ell \in \{ 1, \ldots, t_{k-1} \}}$, subset of timesteps $\mathcal{T} \subseteq \{ 1, \ldots, T \}$ for which we apply the inference model
\ENSURE Approximate sample of the posterior distribution $\posterior{0}{}{}$ defined in \eqref{eq:posterior-distribution}
\STATE $\vect{x}_t \sim \gauss(0, \Id_d)$
\STATE $\vect{x}_0 \gets \denoiser{T}{}{\vect{x}_t}[\theta]$
\FOR{$k = K-1 \to 2$}
    \STATE $s \sim \text{Categorical}((\omega_k^\ell)_{\ell=1, \ldots, t_{k-1}})$
    \STATE $\vect{x}_t \sim \fw{t_k \tbar 0, t_{k+1}}{\vect{x}_0, \vect{x}_t}{}$
    \FOR{$r = 1 \to R$}
        \STATE $\vect{c}\gets (\vect{x}_0, \vect{x}_t, s, t_k, \obs, \operator)$ \label{line:begin-var-approx}
        \STATE $(\vect{\mu}_{\mathrm{zs}}, \vect{\rho}_{\mathrm{zs}}) \gets (\vect{\mu}_{s \tbar 0, t_k}(\vect{x}_0, \vect{x}_t), \sigma^2_{s \tbar 0, t_k} \oned{d})$ \hfill \COMMENT{Zero-shot initialization}
        \IF{$t_k \in \mathcal{T}$}
            \STATE $(\vect{\mu}, \vect{\rho}) \gets \vect{\lambda}_\varphi(\vect{c})$ \hfill \COMMENT{Warm-start candidate}\label{line:inference-model}
            \IF{$\mathcal{L}(\vect{\mu}, \vect{\rho}; \vect{c}) > \mathcal{L}(\vect{\mu}_{\mathrm{zs}}, \vect{\rho}_{\mathrm{zs}}; \vect{c})$}
                \STATE $(\vect{\mu}, \vect{\rho}) \gets (\vect{\mu}_{\mathrm{zs}}, \vect{\rho}_{\mathrm{zs}})$ \hfill \COMMENT{Safeguard fallback}
            \ENDIF
        \ELSE
            \STATE $(\vect{\mu}, \vect{\rho}) \gets (\vect{\mu}_{\mathrm{zs}}, \vect{\rho}_{\mathrm{zs}})$ , cf.~\eqref{eq:prior-mean-init-zs} and \eqref{eq:prior-var-init-zs}\hfill
        \ENDIF
        \FOR{$g = 1 \to G_k$}
            \STATE $(\vect{\mu}, \vect{\rho}) \gets (\vect{\mu}, \vect{\rho}) - \eta \, \nabla_{(\vect{\mu}, \vect{\rho})} \mathcal{L}(\vect{\mu}, \vect{\rho}; \vect{c})$ \label{line:gradient-steps-var-approx}\hfill \COMMENT{Likelihood-guidance}
        \ENDFOR
        \STATE $\vect{x}_s \sim \gauss(\vect{\mu}, \text{diag}(\vect{\rho}))$ \hfill \COMMENT{Sampling $\vect{x}_s$ knowing $\vect{x}_0$ and $\vect{x}_t$}\label{line:end-var-approx}
        \STATE $\vect{x}_0 \gets \text{DDIM}(\vect{x}_s, s, M)$  \hfill \COMMENT{Sampling $\vect{x}_0$ knowing $\vect{x}_s$} \label{line:backward-gibbs}
        \STATE $\vect{x}_t \sim \fw{t_k \tbar s}{\vect{x}_s}{}$   \hfill \COMMENT{Sampling $\vect{x}_t$ knowing $\vect{x}_s$} \label{line:forward-gibbs}
    \ENDFOR
\ENDFOR
\RETURN $\vect{x}_0$
\end{algorithmic}
\end{algorithm}

\section{Details of the Experimental Protocol (\Cref{sec:experiments})}
\label{app:details-experimental-protocol}

This section regroups all the important details to reproduce the results in \Cref{sec:experiments}.

\paragraph{Training, Validation and Test Datasets.}
All experiments are conducted at a resolution of $256 \times 256$, with images resized and center-cropped accordingly.

For FFHQ, the images used for training the unconditional denoiser correspond to indices ranging from $0$ to $59\,999$.
The validation set consists of images with indices from $61\,000$ to $61\,031$, and the test set (used for the final evaluation of all methods) consists of images with indices from $69\,000$ to $69\,299$.

For ImageNet, we use a deterministic subset corresponding to 1\% of the official training split to train the inference model, ensuring that these images are fixed across runs and do not overlap with the validation or test sets.
We use only images from the official validation split to construct the evaluation sets.
We construct two disjoint subsets by selecting at most one image per class: the validation set is formed by taking the first image of each class, while the test set is formed by taking the last image of each class.
We retain 32 images for validation and 300 images for testing.

\paragraph{Normalization of Pixel Values.} To give a meaning to the relative noise level $\sigma_{\obs}$ in \eqref{eq:bayesian-formulation-inv-prob}, we assume that the pixel values are normalized between $-1$ and $1$.

\paragraph{Inference Model Training.} The inference model used in \ours is trained by minimizing the population loss \eqref{eq:inference-model-training} using AdamW \citep{loshchilov2018decoupled}.
We consider two training setups depending on the target dataset.

\begin{itemize}[leftmargin=*, nolistsep, topsep=0pt]
    \item \textbf{FFHQ.} The inference model is trained on the full \texttt{FFHQ} training set (100\%, 60\,000 images) for 5 epochs, with a learning rate of $1 \times 10^{-4}$ and a weight decay of $1 \times 10^{-5}$.
    \item \textbf{ImageNet.} The inference model is trained on 1\% of the ImageNet training set (12\,800 images) for 20 epochs, with a learning rate of $5 \times 10^{-5}$ and no weight decay.
\end{itemize}

In both cases, training is performed with a batch size of 16, using images at a resolution of $256 \times 256$, using a single NVIDIA V100 or H100 GPU for \texttt{FFHQ} and \texttt{ImageNet}, respectively.
The subset of timesteps $\mathcal{T} \subseteq \{ 1, \ldots, T\}$ that is chosen in practice is of the form $\mathcal{T} := \{ 1, \ldots, \lceil (1 - r_{\text{switch}}) T \rceil \}$, for a certain parameter $r_{\text{switch}} \in [0, 1]$ that we tune.

\paragraph{Hyperparameters for \ours and MGDM.}
We describe the hyperparameters of \Cref{alg:inference} (\ours) and MGDM that we consider in our experiments. 
We set the weights $(\omega_t^s)_{s=1}^{t-1}$ of the approximation mixture
$
\posterior{t}{}{\vect{x}_t} \;\approx\; \approxposterior{t}{}{\vect{x}_t}
\;\eqdef\; \sum_{s=1}^{t-1} \omega_t^s \, \approxposterior{t}{}{\vect{x}_t}[s]
$
(see \Cref{sec:background}) to $\omega_t^s = 0$ for $s < t-1$ and $\omega_t^{t-1} = 1$, as this configuration proved most effective in practice.
We always choose a number of $K = 100$ timesteps, and the number of Gibbs repetition is $R = 1$. The timesteps $(t_k)_{k=0}^K$ is a uniform grid between $t_0 = 0$ and $t_K = T$. 
The choice of the subset of timesteps $\mathcal{T} \subseteq \{ 1, \ldots, T \}$ for which we apply the inference model is the same as the one considered during training: $\mathcal{T} := \{ 1, \ldots, \lceil (1 - r_{\text{switch}}) T \rceil \}$, for a certain parameter $r_{\text{switch}} \in [0, 1]$ that we tune.
The number of gradient steps $(G_k)_{k=1}^K$ are chosen as follow:
\begin{equation}
    G_k := \begin{cases}
        G_{\text{start}} & \text{if } t_k \notin \mathcal{T}, \\
        G_{\text{end}} & \text{otherwise},
    \end{cases}
\end{equation}
where $G_{\text{start}}$, $G_{\text{end}}$ are hyperparameters to tune. The number of denoising steps $M$ is tuned as well.
We use the Adam optimizer \citep{kingma2014adam} without weight decay to minimize the variational objective function \eqref{eq:variational-app-gaussian}, with a learning rate $\eta$ to tune. 

\paragraph{Time Measurements.}
All runtimes are measured on a single NVIDIA V100 GPU with a batch size of $1$.
For each image, we measure the wall-clock time of the full sampling loop of the selected method using Python’s high-resolution timer (\texttt{time.perf\_counter}). Timing starts immediately before the sampling call and stops right after it returns; consequently, preprocessing steps (e.g., degradation setup and noise generation), metric computation, and file I/O are excluded.
All experiments are conducted using Python~3.12, PyTorch~2.6, and CUDA~12.4.

\paragraph{Image Reconstruction Tasks.} 
We describe the degradation operators corresponding to the different considered reconstruction tasks: super-resolution, inpainting and motion deblurring. We rely on the implementations of degradation operators provided in the codebase of \citet{janati2025a,elata2025invfussion}.
\begin{itemize}
    \item \textbf{Super-resolution:} The degradation operator is a uniform downsampling operation, by averaging over non-overlapping patches of size $r \times r$. We set $r = 4$, corresponding to the task of $\times 4$ super-resolution.
    \item \textbf{Inpainting:} The degradation operator is a masking operator, where only pixels corresponding to unmasked indices are retained, and the inverse problem is to reconstruct the full image. We consider the following masks. Center inpainting (for \Cref{subsec:acceleration_id} and \Cref{subsec:robustesse_ood}): we remove a central square region covering half of the image ($I/2 \times I/2$ pixels for an image of size $I \times I$). Missing rectangle (for \Cref{subsec:robustesse_ood}): we choose to mask a rectangle of random shape and random position, as implemented by \citet{elata2025invfussion}. Missing pixels (for \Cref{subsec:robustesse_ood}): we choose to mask some randomly selected pixels, following the implementation of MissingPatch in \citet{elata2025invfussion}.
    \item \textbf{Motion deblurring:} The degradation operator is a convolution with a fixed motion blurring kernel. We use a kernel of size $21 \times 21$ with intensity $0.9$ in \Cref{subsec:acceleration_id}, following the implementation of MotionBlur in \citet{janati2025a}. We also use kernel of size $35 \times 35$ with intensity $0.3$ in \Cref{subsec:robustesse_ood}.
\end{itemize}

\subsection{Protocol for the Inference Time Acceleration with Amortization (\Cref{subsec:acceleration_id})}
\label{app:acceleration}

For the MGDM baseline, we perform an extensive hyperparameter search on 32 validation images, evaluating more than 90 combinations to determine the Pareto front of reconstruction quality (measured with LPIPS, a perceptual similarity metric well aligned with human visual judgments) versus inference time.
This Pareto front represents the set of configurations that yield the best trade-off between reconstruction fidelity and inference speed for MGDM.
Using the same hyperparameter grid, we tune \ours to identify configurations surpassing this Pareto front.

\paragraph{Hyperparameter Grid.}
To characterize the trade-off between reconstruction quality and inference speed for both MGDM and \ours, we evaluate a broad grid of hyperparameter configurations.
We adopt the naming conventions introduced in \Cref{alg:inference}.
The grid comprises:
\begin{itemize}
    \item $G_{start} \in \{1, 3\}$  
    \item $G_{end} \in \{0, 1, 3, 10\}$  
    \item Learning rate $\eta \in \{0.01, 0.03\}$  
    \item Number of denoising steps $M \in \{1, 5\}$  
    \item Switch ratio $r_{\text{switch}} \in \{70\%, 80\%, 90\%\}$  
\end{itemize}

\paragraph{Pareto Front Construction.}
We trace the MGDM Pareto front by sweeping the hyperparameters presented previously on a 32-image validation set, thereby characterizing the trade-off between reconstruction quality and inference time. We then take the best zero-shot configurations (highest quality at a given speed) and evaluate it on 300 images from the test set.

\paragraph{Statistical Test.}
We compare \ours to the zero-shot baseline using a paired non-inferiority \(t\)-test on per-image LPIPS scores (denoted by $\mathrm{LPIPS}_{\mathrm{ZS}}$ for the zero-shot baseline, and $\mathrm{LPIPS}_{\mathrm{AM}}$ for \ours, lower is better). For each test image $i$, we form the paired difference
\begin{equation}
  \delta_i := \mathrm{LPIPS}_{\mathrm{ZS},i} - \mathrm{LPIPS}_{\mathrm{AM},i},
\end{equation}
so that larger $\delta_i$ favors \ours.

With non-inferiority margin \(\Delta>0\), the hypotheses are
\begin{align}
  H_0 &: \ \delta \le -\Delta 
  && \text{(\ours worse than MGDM by at least \(\Delta\))}, \\
  H_1 &: \ \delta >  -\Delta 
  && \text{(\ours is not inferior to MGDM)}.
\end{align}

Let $\bar \delta$ and $\bar{s}$ be the empirical mean and standard deviation of the samples $\{\delta_i\}_{i=1}^n$. The statistical test is
\begin{equation}
  T \;=\; \frac{\bar \delta + \Delta}{\,\bar{\delta}/\sqrt{n}\,}, 
\end{equation}
We reject \(H_0\) at level \(\alpha\) if
\begin{equation}
  T > t_{1-\alpha,\,n-1},
\end{equation}

where $t_{1-\alpha,\,n-1}$ denotes the $(1-\alpha)$-quantile of the Student $t$ distribution with $n-1$ degrees of freedom. Equivalently, we conclude non-inferiority if the one-sided \((1-\alpha)\) lower confidence bound for \(\delta\) exceeds \(-\Delta\).

We use \(\alpha=0.05\) and  $\Delta=0.003$. Configurations of \ours that reject \(H_0\) are considered to be configurations that does not degrade the reconstruction quality compared to the considered  MGDM sampling. They are therefore retained for speedup reporting when aiming for a certain target quality, as it is done in \Cref{tab:speedup_results_compact}.

\subsection{Protocol for Out-of-distribution Robustness (\Cref{subsec:robustesse_ood})}

The architecture of the inference model for the inpainting task in \Cref{subsec:robustesse_ood} is based on Palette \cite{saharia2022palette}, and the one for the motion deblurring task is based on InvFussion \cite{elata2025invfussion}.

Given a reconstruction task and a fixed inference computational budget, each method (\ours and MGDM) is tuned over the same hyperparameter grid described in \Cref{app:acceleration}. The optimal hyperparameter configuration is selected on a validation set of 32 images by maximizing a chosen quality metric (e.g., LPIPS or CMMD). Final performance is then reported on a held-out test set of 300 images. 

\section{Additional Experiments}
\label{app:additional_experiments}

This section contains additional experiments to reinforce the motivation of this paper and to further support our claims.

\subsection{Lack of Out-of-distribution Robustness for Supervised Diffusion}
\label{app:lack_robustness_ood_sft}

Our method relies on explicit likelihood guidance, as in posterior sampling, which we show to provide an advantage in robustness to out-of-distribution operators compared to InvFussion \citep{elata2025invfussion}—a supervised diffusion baseline that embeds the degradation operator into the conditional denoiser.
While this design is effective when degradations match those seen during training, we hypothesize that it struggles to generalize when the inference-time operator deviates from this distribution.
In contrast, likelihood guided methods enforces data consistency with the actual operator at inference time, thereby avoiding this generalization bottleneck.

To test our hypothesis, we reproduced the supervised baseline InvFussion  \citep{elata2025invfussion} following its exact setup for image inpainting, considering no more a fixed mask, but different families of masks during training and inference. During training, the learned conditional denoiser only observes masks that remove a single randomly placed rectangle (in-distribution degradations). 
At evaluation, we introduce a different family of masks in which individual pixels are randomly removed according to an i.i.d.~Bernoulli sampling with parameter \(p \in [0.9, 0.92]\), creating out-of-distribution degradations. In order to provide a reference, we evaluate MGDM on the same tasks.

\paragraph{Training and Inference of InvFussion.}
The training of the supervised baseline InvFussion \citep{elata2025invfussion} uses the full training set of \texttt{FFHQ} at resolution $64 \times 64$ (60000 images), following the exact protocol of \citet{elata2025invfussion}. In their protocol, they use an Adam optimizer without weight decay, with a reference learning rate of $5 \times 10^{-5}$ that follows a warm-up-and-decaying schedule. The model is trained on $32768$ kimgs with a batch size of 512.
At inference, we use the DDIM sampler with $500$ steps to generate reconstructions with the supervised diffusion baseline InvFussion \citep{elata2025invfussion}.
For comparison of the reconstruction quality, we compare with the zero-shot posterior sampling baseline MGDM \citep{janati2025a}.
Its hyperparameters are the following: we use $K=100$ diffusion steps, the number of gradient steps are $G_{\text{start}} = 3$ and $G_{\text{end}} = G$ with $G \in \{0, 30 \}$ gradient updates applied to the variational approximation during the final 10 steps, \emph{i.e}, for timesteps $t_k$ with $k \in \{ 1, \ldots, 10 \}$. The learning rate $\eta$ for the variational approximation problem is set to $0.01$. The number of Gibbs repetition is $R=1$, and the number of denoising steps is $M=1$. 

\paragraph{Results.}

\Cref{tab:ood-masks} shows that the supervised baseline InvFussion \citep{elata2025invfussion} performs well on in-distribution masks but collapses under out-of-distribution masks, despite explicitly encoding the degradation operator. See \Cref{fig:figure-intro} for an illustration: the reconstruction of InvFussion on out-of-distribution masks has artefacts. This illustrates its lack of robustness when the inference-time degradation departs from what was modeled during training.
In contrast, MGDM maintains strong reconstruction quality across both mask types. This is because MGDM is a zero-shot method whose performance is driven by the test-time guidance term, which ensures that the generated samples remain consistent with the actual observation regardless of the specific mask geometry.

These findings highlight the central role of test-time likelihood guidance in achieving robustness to out-of-distribution degradations. They also illustrate the limitations of supervised baselines, which fail when the degradation operator is unseen during training. 

\begin{table}[h]
\centering
\scriptsize
\setlength{\tabcolsep}{4pt}
\caption{Supervised diffusion InvFussion  \citep{elata2025invfussion} vs.~MGDM \citep{janati2025a} on inpainting with out-of-distribution masks (missing pixels). Evaluation on 300 images of \ffhq. Models were trained on in-distribution masks (single missing rectangle). The best metric is in bold.
}
\label{tab:ood-masks}
\resizebox{0.80\columnwidth}{!}{%
\begin{tabular}{lc c c c}
\toprule
& LPIPS $\downarrow$ & SSIM $\uparrow$ & PSNR $\uparrow$ & CMMD $\downarrow$ \\
\midrule
In-distribution masks (missing rectangle) \\
\midrule
InvFussion \citep{elata2025invfussion} & 0.015 $\pm$ 0.027 & 0.947 $\pm$ 0.075 & 33.2 $\pm$ 8.1 & \textbf{0.010} \\
MGDM \citep{janati2025a}      & \textbf{0.014 $\pm$ 0.024} & \textbf{0.952 $\pm$ 0.066} & \textbf{33.3 $\pm$ 7.1} & 0.011 \\
\midrule
Out-of-distribution masks (missing pixels) \\
\midrule
InvFussion \citep{elata2025invfussion} & 0.403 $\pm$ 0.118 & 0.427 $\pm$ 0.077 & 16.8 $\pm$ 1.4 & 1.029 \\
MGDM \citep{janati2025a}      & \textbf{0.040 $\pm$ 0.022} & \textbf{0.846 $\pm$ 0.051} & \textbf{23.6 $\pm$ 2.5} & \textbf{0.077} \\
\bottomrule
\end{tabular}
}%
\end{table}

\subsection{Comparison with Other Zero-shot Posterior Sampling Methods}
\label{app:ffhq256_dps}

We put into perspective the trade-off between reconstruction quality and inference time of MGDM studied in \Cref{subsec:acceleration_id} compared to other existing zero-shot posterior sampling algorithm, such as DPS \citep{chung2023diffusion}.

\paragraph{Protocol.}
The zero-shot DPS baseline is tuned on the validation set (32 images). We vary the number of sampling steps ($T \in {10, 30, 100, 300, 1000}$) and the step sizes ($\zeta \in {0.1, 0.3, 1, 3}$), and retain for each $T$ the best-performing configuration. The final evaluation is conducted on the test set (300 images). This procedure yields the reconstruction–time trade-off for DPS, which we compare against MGDM trade-off studied in \Cref{subsec:acceleration_id}.

\begin{figure}[h]
  \centering
  \includegraphics[width=0.4\linewidth]{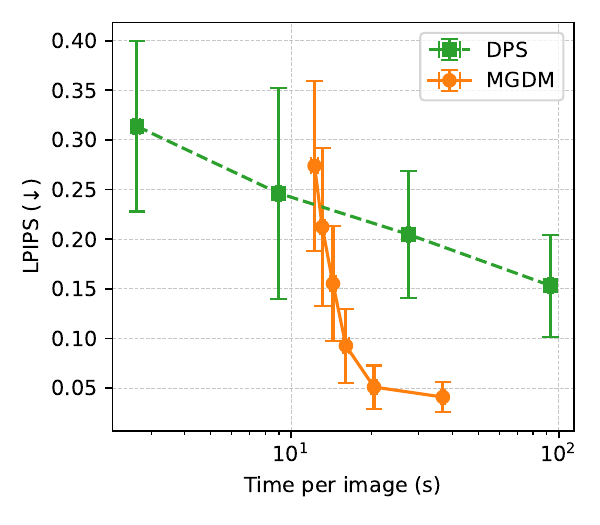}
  \caption{Trade-off between reconstruction quality (LPIPS) and inference time of MGDM \citep{janati2025a} vs.~DPS \citep{chung2023diffusion}, for motion deblurring in \texttt{FFHQ}. Configurations of MGDM and DPS are Pareto-optimal.}
  \label{fig:dps_vs_mgdm_lpips_vs_time}
\end{figure}

\paragraph{Results.}
Figure~\ref{fig:dps_vs_mgdm_lpips_vs_time} shows that MGDM consistently achieves a better trade-off between reconstruction quality and inference time than DPS. Across the full range of tested compute budgets, MGDM obtains lower LPIPS at equal inference time. For example, for inference times below 30 seconds, DPS reaches an average LPIPS of $0.205$, whereas MGDM attains a substantially lower LPIPS of $0.044$. Even when allowing longer runtimes, DPS only improves to an average LPIPS of $0.153$ at 98 seconds of inference.
These results are consistent with prior observations that MGDM provides higher reconstruction fidelity than other zero-shot sampling methods~\citep{janati2025a}.

\subsection{Training Cost vs. Inference Efficiency Analysis in \ours}
\label{app:ffhq256_computational_cost}

We analyze the computational investment required to train the inference model in \ours compared to the MGDM baseline. We demonstrate that the training overhead for the amortization module is modest and is rapidly compensated by the significant acceleration achieved during inference.

\paragraph{Protocol.}
We compare MGDM and \ours under configurations selected to achieve comparable high-fidelity reconstruction quality (LPIPS $\approx 0.045$). 
\begin{itemize}[leftmargin=*, nolistsep]
    \item \textbf{MGDM:} We use inference hyperparameters $G_\text{end} = 10$, $G_\text{start} = 1$, $M = 1$, learning rate $\eta = 0.03$, and a switch ratio $r_\text{switch} = 80\%$.
    \item \textbf{\ours:} We utilize the same inference settings but increase the switch ratio to $r_\text{switch} = 90\%$ to leverage the amortized initialization. The inference model is trained on $10\%$ of the \texttt{FFHQ} dataset.
\end{itemize}

\begin{table}[h]
\centering
\caption{
Training time, inference time (average) and LPIPS (average and standard error) for MGDM and \ours on motion deblurring in \texttt{FFHQ}. Metrics are computed over $300$ test images.
}
\resizebox{0.70\textwidth}{!}{%
\begin{tabular}{l c c c}
\toprule
Method & Training time (GPUh) & Inference time / image (s) & LPIPS $\downarrow$ \\
\midrule
MGDM        & $0$      & $28.7$             & $0.044 \pm 0.018$ \\
\ours        & $6.6$   & $20.5$             & $0.045 \pm 0.017$ \\
\bottomrule
\end{tabular}%
}
\label{tab:endtoend_compute}
\end{table}

\paragraph{Results.}
As reported in \Cref{tab:endtoend_compute}, training the inference model for \ours requires approximately $6.6$ GPU-hours. Although MGDM has no initial training cost, it is slower at inference ($28.7$ s/image) compared to \ours ($20.5$ s/image). By defining the total cost as the sum of training time and cumulative inference time, we find that the initial training overhead is fully amortized after approximately $N = 2{,}900$ images. Consequently, for any workload exceeding this relatively small number of reconstructions, \ours is strictly more computationally efficient than the zero-shot baseline.

\subsection{Extended Pareto Frontier Analysis on \texttt{FFHQ}}
\label{app:additional_acceleration_id_pareto_ffhq}

\begin{figure*}[h]
\centering
\begin{subfigure}[t]{0.40\linewidth}
\centering
\includegraphics[width=\linewidth]{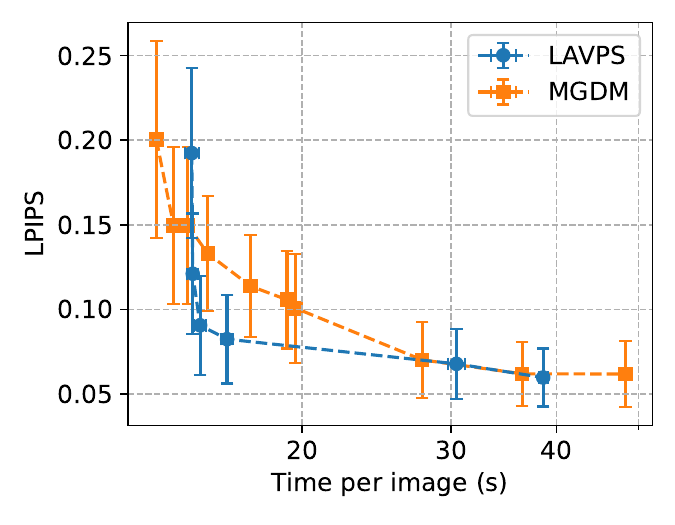}
\caption{Motion deblurring}
\label{fig:ffhq-pareto-mb21}
\end{subfigure}
\begin{subfigure}[t]{0.40\linewidth}
\centering
\includegraphics[width=\linewidth]{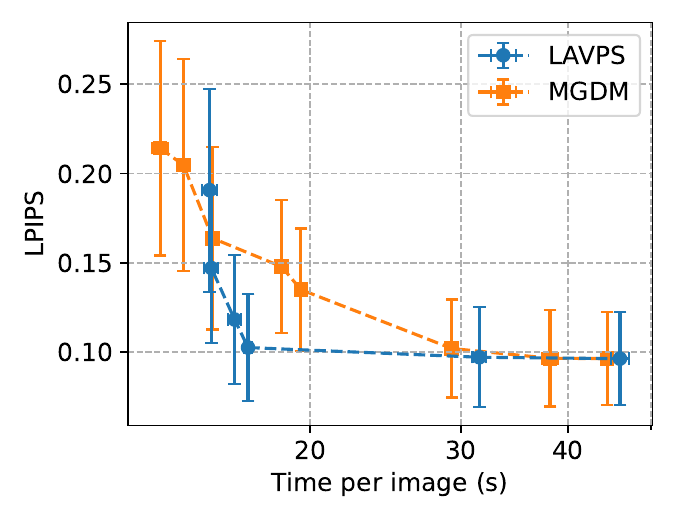}
\caption{Super-resolution $\times 4$}
\label{fig:ffhq-pareto-sr4}
\end{subfigure}
\caption{Trade-off between reconstruction quality and inference time of \ours vs.~MGDM on the \texttt{FFHQ} dataset. The configurations are Pareto-optimal.}
\label{fig:ffhq-pareto}
\end{figure*}

Under the same protocol as in \Cref{subsec:acceleration_id}, \Cref{fig:ffhq-pareto} expand our Pareto frontier analysis in \Cref{subsec:acceleration_id} to include reconstruction tasks on the \texttt{FFHQ} dataset, such as motion deblurring and super-resolution. This further validates the improved trade-off between reconstruction quality and inference speed for \ours compared to MGDM. 

\subsection{Ablation Study: Impact of Gradient Steps on Convergence}
\label{app:additional_acceleration_id_gradient_steps}

We analyze how the number of inference gradient steps, denoted $G_{\text{end}}$, affects both reconstruction quality and inference speed for \ours and MGDM. This comparison highlights the efficiency of the likelihood-aware warm-start.

\paragraph{Protocol.}
MGDM hyperparameters are fixed as follows: switch ratio $r_{\text{switch}} = 80\%$; the number of gradient steps for timesteps outside $\mathcal{T} := \{1, \ldots, \lceil (1 - r_{\text{switch}}) T \rceil \}$ is $G_{\text{start}} = 1$; the learning rate is $\eta = 0.03$; and the number of denoising steps is $M=1$.
We then vary the number of gradient steps $G_{\text{end}} \in \{0, 1, 3, 10\}$ timesteps in $\mathcal{T}$, to study its role in both LVAPS and MGDM.
We considered here $32$ images from the validation set.

\begin{figure}[h]
\centering
\begin{tabular}{cc}
    \parbox{0.40\linewidth}{
        \centering
        \includegraphics[width=\linewidth]{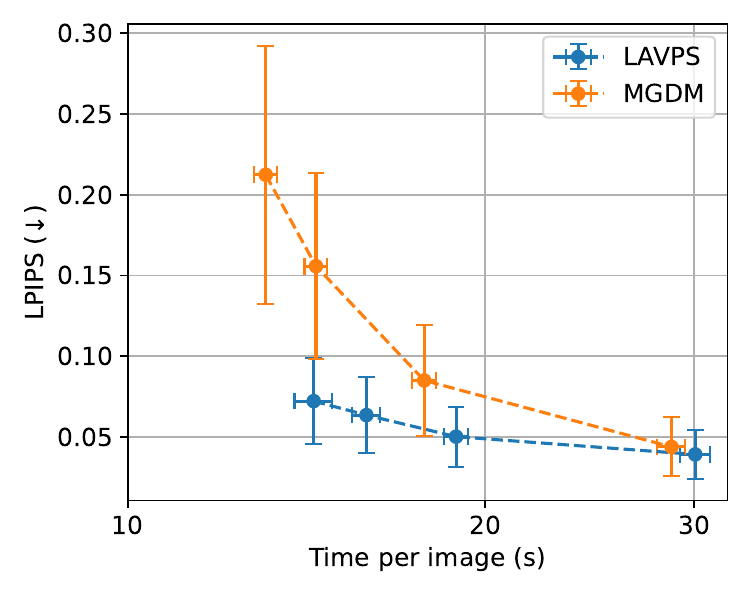}
    }
    &
    \parbox{0.40\linewidth}{
        \centering
        \resizebox{\linewidth}{!}{%
            \begin{tabular}{rcc}
                \toprule
                $G_\text{end}$ & MGDM & \ours \\
                \midrule
                0  & $0.212 \pm 0.080$ & $0.072 \pm 0.027$ \\
                1  & $0.156 \pm 0.057$ & $0.063 \pm 0.023$ \\
                3  & $0.085 \pm 0.034$ & $0.050 \pm 0.019$ \\
                10 & $0.044 \pm 0.018$ & $0.039 \pm 0.015$ \\
                \bottomrule
            \end{tabular}%
        }
    }
\end{tabular}

\caption{Impact of the number of gradient steps $G_\text{end}$ on reconstruction performance and inference time for motion deblurring on \texttt{FFHQ}. Left: LPIPS vs.\ inference time for \ours and MGDM. Right: corresponding LPIPS values (mean $\pm$ std) for the different values of number of gradient steps $G_\text{end}$.}
\label{fig:ablation-grad-steps}
\end{figure}

\paragraph{Results.} As shown in \Cref{fig:ablation-grad-steps}, \emph{for the same choice of hyperparameter}, \ours is slightly slower than MGDM due to the additional forward pass through the inference model used to warm-start the variational problem. However, this small overhead is outweighed by a substantial improvement in reconstruction quality, confirmed quantitatively in \Cref{fig:ablation-grad-steps}.
Overall, \ours maintains high reconstruction quality even with very few gradient steps, whereas MGDM performs poorly in the regime with a low number of gradient steps. This highlights that the learned inference model enables \ours to operate with significantly fewer gradient steps (hence, a lower inference time) while preserving performance, leading to a better quality-time trade-off.

\subsection{Pareto Frontier Analysis When Amortizing on a Family of Operators}
\label{app:additional_exp_section_5p2}

\begin{figure*}[h]
    \centering
    \begin{subfigure}[t]{0.48\linewidth}
        \centering
        \includegraphics[width=\linewidth]{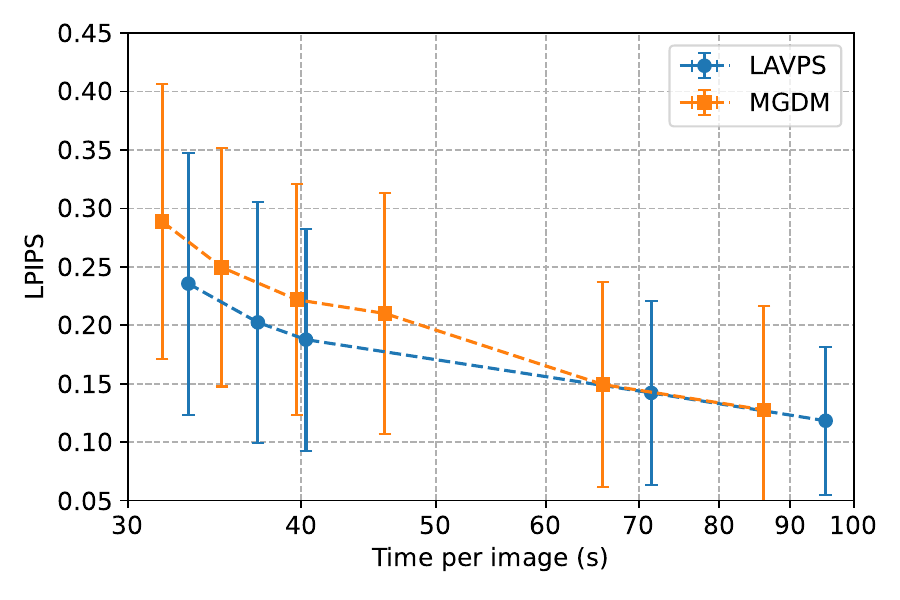}
        \caption{Motion deblurring (kernel $21 \times 21$)}
        \label{fig:mb21}
    \end{subfigure}
    \begin{subfigure}[t]{0.48\linewidth}
        \centering
        \includegraphics[width=\linewidth]{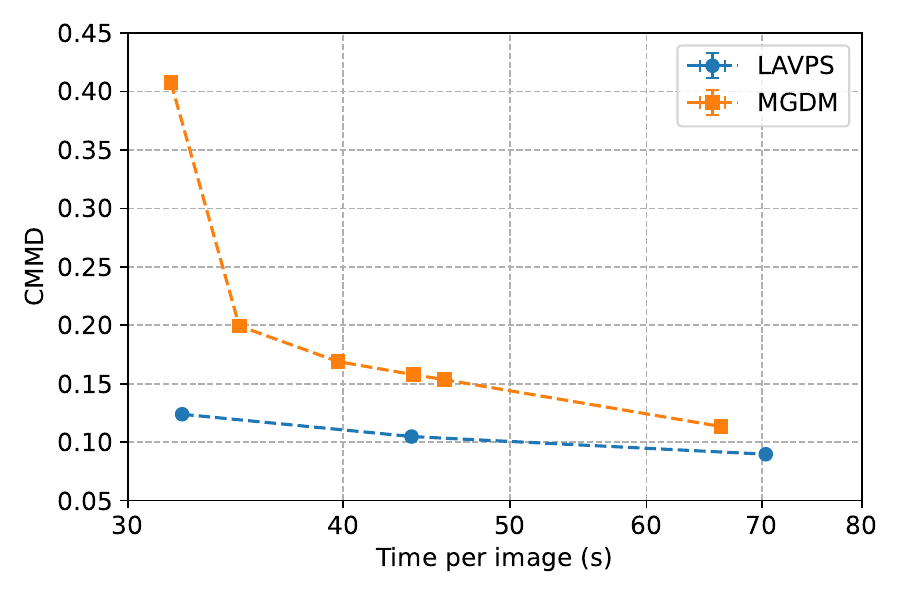}
        \caption{Inpainting (missing rectangle)}
        \label{fig:mb35}
    \end{subfigure}

    \caption{Efficiency-quality trade-off for \ours vs. MGDM, {on in-distribution degradation operators}. The inference model in \ours is trained on a family of degradation operators. Markers represent Pareto-optimal configurations for each method across various restoration tasks on the \texttt{Imagenet} dataset.}
    \label{fig:imagenet256_motiondeblurring_ood}
\end{figure*}

\Cref{fig:imagenet256_motiondeblurring_ood} shows that, for the reconstruction tasks considered in \Cref{subsec:robustesse_ood}, the improved trade-off between reconstruction quality and inference speed achieved by \ours over MGDM persists even when the inference model is trained on a family of operators. The protocol used to construct the Pareto front is identical to that of \Cref{subsec:acceleration_id}, except that the inference model is trained on a family of operators rather than a single operator.

\begin{figure*}[h]
    \centering
    \begin{subfigure}[t]{0.48\textwidth}
        \centering
        \includegraphics[width=\linewidth]{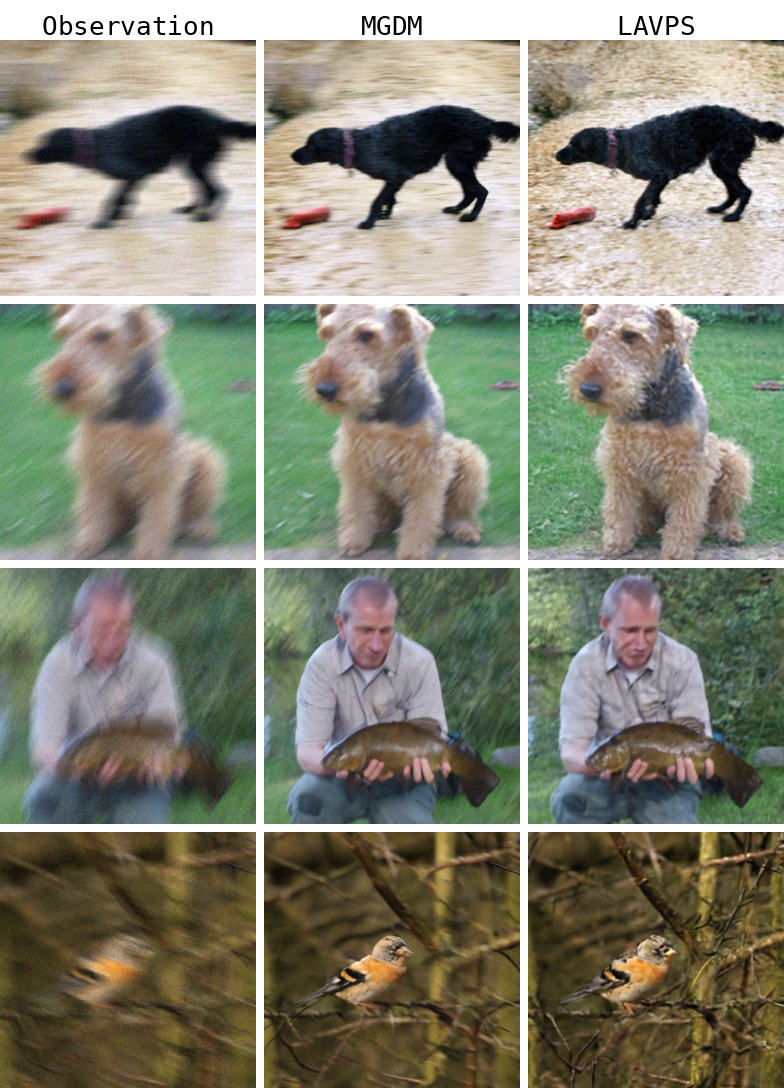}
        \caption{\textbf{In-distribution (kernel size $21 \times 21$)}. MGDM is blurry on the background and the animal.}
    \end{subfigure}
    \hfill
    \begin{subfigure}[t]{0.48\textwidth}
        \centering
        \includegraphics[width=\linewidth]{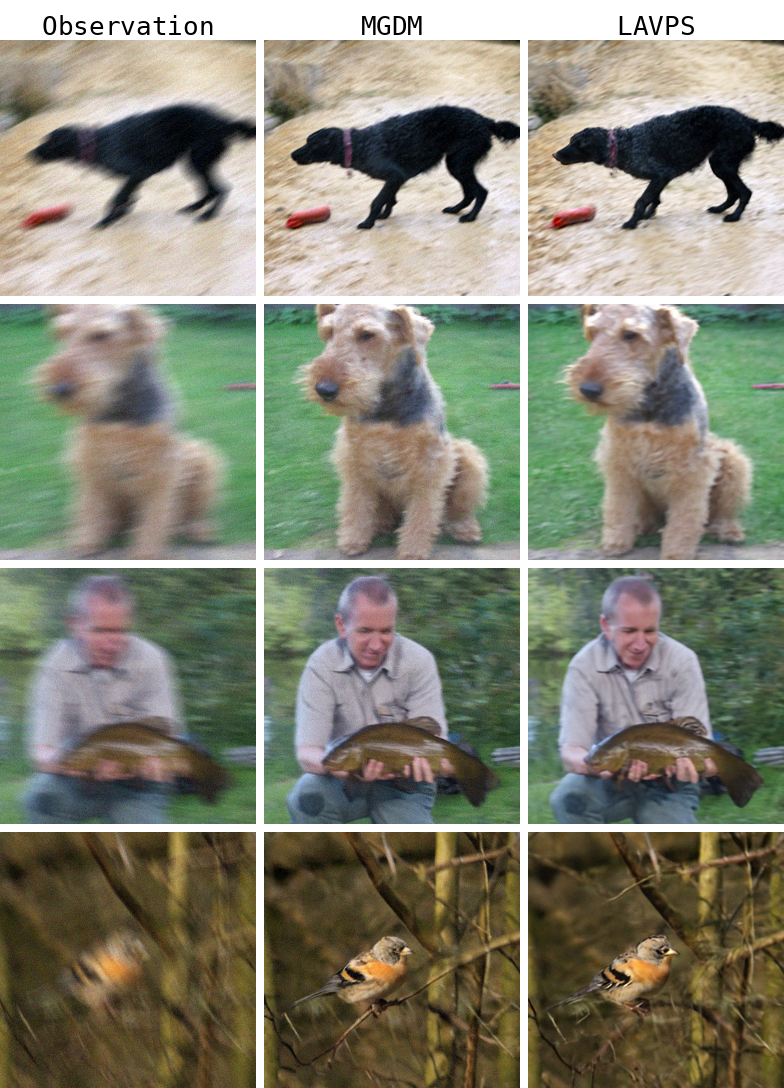}
        \caption{\textbf{Out-of-distribution (kernel size $35 \times 35$)}. \ours does not degrade compared to MGDM.}
    \end{subfigure}
    \caption{Motion deblurring on \texttt{ImageNet}. 
    The inference time is constrained to be lower than $47.3$s.}
    \label{fig:mb_qualitative_47p3_appendix}
\end{figure*}

\begin{figure*}[h]
    \centering
    \begin{subfigure}[t]{0.48\textwidth}
        \centering
        \includegraphics[width=\linewidth]{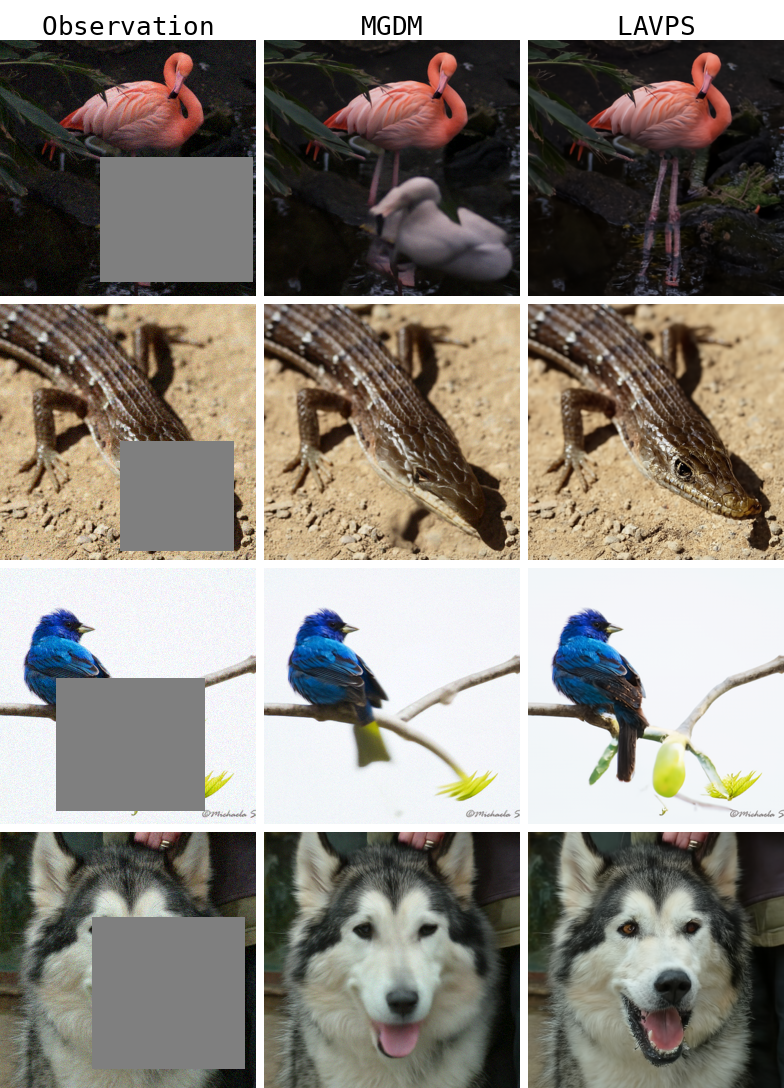}
        \caption{\textbf{In-distribution (missing rectangle)}. MGDM produces hallucinated content or yields blurry reconstruction.}
    \end{subfigure}
    \hfill
    \begin{subfigure}[t]{0.48\textwidth}
        \centering
        \includegraphics[width=\linewidth]{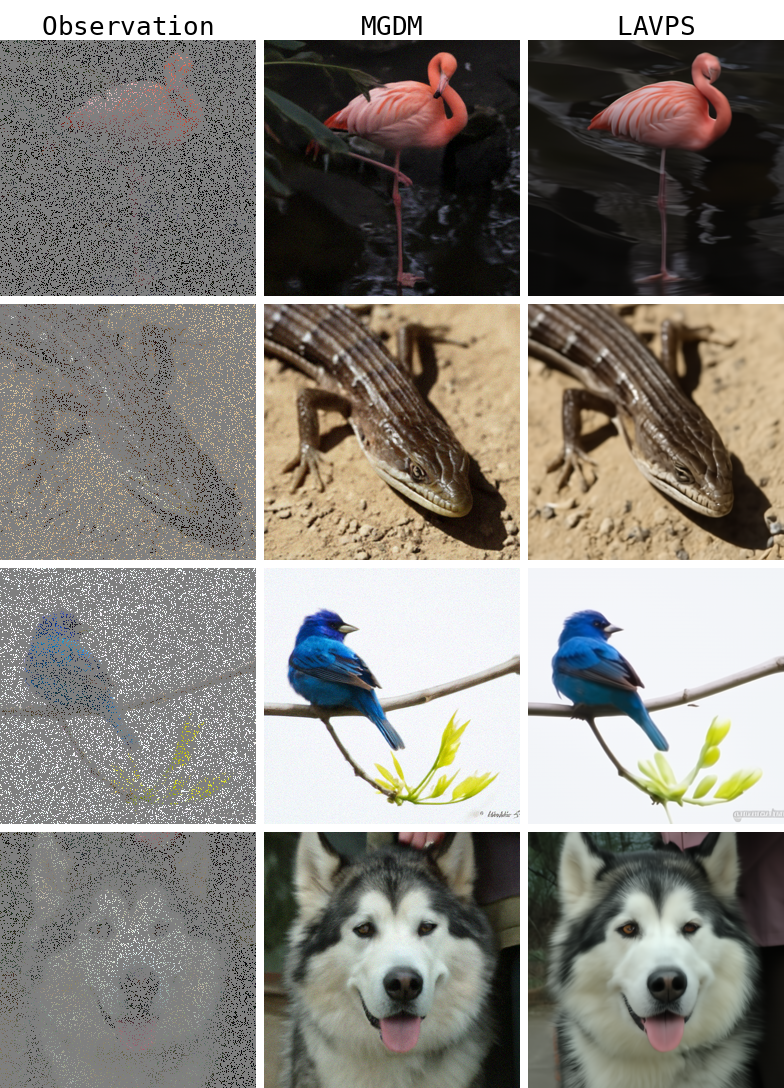}
        \caption{\textbf{Out-of-distribution (missing pixel)}. \ours does not degrade compared to MGDM.}
    \end{subfigure}
    \caption{
    Inpainting on \texttt{ImageNet}. 
    The inference time is constrained to be lower than $34.9$s.
    }
    \label{fig:inpainting_qualitative_34p9_appendix}
\end{figure*}

As claimed in \Cref{subsec:robustesse_ood}, this improvement on in-distribution opeartors does not sacrifice the flexibility on arbitrary opeartors of posterior sampling methods: \ours remains robust to out-of-distribution operators. We complement \Cref{fig:mb_qualitative_47p3,fig:inpainting_qualitative_34p9} with additional reconstruction samples in \Cref{fig:mb_qualitative_47p3_appendix,fig:inpainting_qualitative_34p9_appendix}. Under a certain target computational budget at inference, \ours achieves better reconstructions than MGDM, e.g., reconstructions are sharper and hallucinate less, while being able to recover the reconstruction quality of MGDM on out-of-distribution operators.

\end{document}